\documentclass[letterpaper]{article} % DO NOT CHANGE THIS
\usepackage{aaai25}  % DO NOT CHANGE THIS
\usepackage{times}  % DO NOT CHANGE THIS
\usepackage{helvet}  % DO NOT CHANGE THIS
\usepackage{courier}  % DO NOT CHANGE THIS
\usepackage[hyphens]{url}  % DO NOT CHANGE THIS
\usepackage{graphicx} % DO NOT CHANGE THIS
\usepackage{natbib}  % DO NOT CHANGE THIS AND DO NOT ADD ANY OPTIONS TO IT
\usepackage{caption} % DO NOT CHANGE THIS AND DO NOT ADD ANY OPTIONS TO IT
\usepackage{algorithm}
\usepackage{algorithmic}
\usepackage{newfloat}
\usepackage{listings}
\DeclareCaptionStyle{ruled}{labelfont=normalfont,labelsep=colon,strut=off} % DO NOT CHANGE THIS
\floatstyle{ruled}
\newfloat{listing}{tb}{lst}{}
\floatname{listing}{Listing}
\usepackage{amsmath}
\usepackage{graphicx}

\usepackage{subcaption}
\title{AdvIRL: Reinforcement Learning-Based Adversarial Attacks on 3D NeRF Models}

\author{%
  Tommy Nguyen \quad Mehmet Ergezer\thanks{Dr. Ergezer holds concurrent appointments as an Associate Professor at Wentworth Institute of Technology and as an Amazon Visiting Academic. This paper describes work performed at Wentworth Institute of Technology and is not associated with Amazon.}  \quad Christian Green}
  \affiliations{
  School of Computing and Data Science\\
  Wentworth Institute of Technology\\
  Boston, MA  \\
  \texttt{\{nguyent68, ergezerm, greenc10\}@wit.edu} \\
}

\usepackage{amssymb}

\begin{document}

\maketitle

\begin{abstract}
  
  %ME suggestion:
  The increasing deployment of AI models in critical applications has exposed them to significant risks from adversarial attacks. While adversarial vulnerabilities in 2D vision models have been extensively studied, the threat landscape for 3D generative models, such as Neural Radiance Fields (NeRF), remains underexplored. This work introduces \textit{AdvIRL}, a novel framework for crafting adversarial NeRF models using Instant Neural Graphics Primitives (Instant-NGP) and Reinforcement Learning. Unlike prior methods, \textit{AdvIRL} generates adversarial noise that remains robust under diverse 3D transformations, including rotations and scaling, enabling effective black-box attacks in real-world scenarios. Our approach is validated across a wide range of scenes, from small objects (e.g., bananas) to large environments (e.g., lighthouses). Notably, targeted attacks achieved high-confidence misclassifications, such as labeling a banana as a slug and a truck as a cannon, demonstrating the practical risks posed by adversarial NeRFs. Beyond attacking, \textit{AdvIRL}-generated adversarial models can serve as adversarial training data to enhance the robustness of vision systems. The implementation of \textit{AdvIRL} is publicly available at \url{https://github.com/Tommy-Nguyen-cpu/AdvIRL/tree/MultiView-Clean}, ensuring reproducibility and facilitating future research.

\end{abstract}

\section{Introduction}
% ME: streamlining flow to emphasize significance and highlight novelty earlier. 

The widespread integration of artificial intelligence (AI) across fields such as autonomous driving and robotics \cite{li2019stereo, ravindran2020multi,zhang2020dnn, hu20193, hossain2016object, finn2015learning} has led to an increase in incidents where AI models misclassify objects \cite{kurakin2017adversarial}. These errors can have severe consequences, including the loss of life, particularly in safety-critical systems like AI-driven vehicles. To mitigate such risks, research on adversarial AI has expanded, aiming to expose vulnerabilities in AI algorithms and strengthen their robustness against adversarial attacks \citep{xu2020adversarial, behzadan2017vulnerability, goodfellow2015explaining}. 

While adversarial attacks on 2D vision models have been widely studied \citep{li2023adv3d, huang2023transferable, madry2017towards}, the transition to the physical world introduces new challenges. Many real-world attacks rely on 3D transformations—such as changes in angle, scale, and lighting—which 2D approaches cannot adequately capture. Further, many existing approaches assume white-box access to the model's architecture and gradients \citep{sarkar2023rlab, lin2020nesterov, andriushchenko2020understanding, dong2018boosting, hull2024revamp}. In contrast, real-world adversarial attacks are often black-box in nature, where attackers do not have direct access to the model but instead exploit vulnerabilities through inputs captured by external devices like cameras \citep{athalye2018synthesizing, Sharif_2019, UniversalPerturb2024}.

Recent efforts have begun exploring adversarial attacks on 3D objects \citep{li2023adv3d}, but this domain remains underdeveloped. Applying adversarial noise to 3D objects is more resource-intensive, often requiring high-quality 3D models and complex manipulation of textures. Moreover, existing datasets for 3D adversarial research are sparse compared to their 2D counterparts, creating a bottleneck for advancements. Although recent studies have explored the 3D adversarial landscape using rendering techniques such as Gaussian Splatting \cite{zeybey2024gaussian}, these investigations remain largely limited to white-box settings. AdvIRL innovatively overcomes these limitations, relying solely on the input and output of the target model to generate effective adversarial attacks. This gradient-free approach simplifies the process and sidesteps challenges associated with adversarial transferability in existing methods.

%ME?
In the context of this research, we focus on \textbf{digital adversarial attacks} targeting machine learning vision systems. These attacks simulate how carefully crafted perturbations can manipulate neural rendering models, potentially compromising perception in critical applications like autonomous vehicles, surveillance systems, and robotic navigation.

\textbf{Practical Scenario Example:} Consider a virtual reality (VR) training application for first responders, where environments like urban landscapes or rescue scenes are reconstructed using  Neural Radiance Fields (NeRFs). The system must render objects accurately from various angles and distances to ensure realism and usability. An adversarial attack on such a system could introduce imperceptible yet impactful distortions, disrupting the visual consistency and undermining the training's effectiveness. Our AdvIRL framework addresses this challenge by generating robust adversarial noise that maintains its efficacy across diverse transformations, allowing for more consistent 3D renderings that minimize distortions and visual inconsistencies when using the adversarial noise generated by AdvIRL during training. 

AdvIRL can also be used to enhance the security of 3D object detection algorithms used in various applications, such as security cameras and advanced tolling systems. By training these algorithms on adversarial noise, they become more resilient to both intentional (perturbations) and unintentional attacks (adversarial rotation). This can help protect sensitive information and improve the accuracy of these systems.
% \#TODO??? Not mandatory; see the example added to Chris' intro: "Consider a security camera system monitoring a 3D object from multiple angles. Traditional universal perturbations might require different adversarial patterns for each view, potentially breaking the attack's effectiveness as the camera angle changes. Our method addresses this limitation by generating a single perturbation that remains robust across viewpoints, making it more practical for real-world adversarial attacks on 3D object recognition systems." I'd avoid self-driving cars, but up to you. 

In this work, we propose \textit{AdvIRL}, a novel black-box reinforcement learning framework that generates adversarial 3D models using a state-of-the-art NeRF architecture \citep{mildenhall2020nerf}. Unlike previous approaches that assume model access, \textit{AdvIRL} operates in a black-box setting, leveraging Instant Neural Graphics Primitives (Instant-NGP) \citep{mueller2022instant} to introduce adversarial noise resilient to diverse transformations. Furthermore, by incorporating image segmentation \citep{wu2019detectron2}, our method can target specific objects within a scene, allowing for more precise and controlled adversarial attacks. Our approach can be utilized to significantly improve the robustness of vision models when adversarial NeRF models are used for adversarial training.

\iffalse
% \textbf{Contributions} 
Our work contributes the following:
\begin{enumerate}
    \item We propose AdvIRL, a novel framework for generating adversarial 3D NeRF models by leveraging Instant-NGP and Detectron2, enabling the generation of robust adversarial noise across various angles and distances. These models can be used to improve the robustness of vision systems against adversarial attacks.
    \item We demonstrate the efficacy of our approach across diverse scenes, from small objects (e.g., a banana) to large environments (e.g., a lighthouse), showcasing its versatility.
    \item To the best of our knowledge, AdvIRL is the first black-box adversarial framework for targeting 3D vision models, combining reinforcement learning with Neural Radiance Fields to generate adversarial images from multiple angles and 3D NeRF models. AdvIRL streamlines the adversarial generation process by requiring only input images and output labels from a target model to produce robust adversarial results.
\end{enumerate}
\fi

This work introduces an innovative approach to black-box adversarial robustness for 3D vision models, with the following key contributions:

\begin{enumerate}
    \item We propose \textit{AdvIRL}, the first black-box adversarial framework for Neural Radiance Fields (NeRF). Our approach leverages reinforcement learning and Instant Neural Graphics Primitives (Instant-NGP) to efficiently generate robust adversarial noise under diverse 3D transformations, such as rotations, scaling, and viewpoint changes in AI vision models. To our knowledge, \textit{AdvIRL} is the first black-box adversarial framework for targeting 3D vision models.
    \item We demonstrate the efficacy of \textit{AdvIRL} across a wide range of scenes, from small-scale objects (e.g., transforming a banana's classification) to large environmental scenes (e.g., manipulating lighthouse representations). Our targeted attacks achieved high-confidence misclassifications (e.g., labeling a banana as a slug and a truck as a cannon), highlighting the practical risks posed by adversarial NeRFs. %Despite advances in training robust vision models, our work shows that these models continue to encounter many of the same challenges they faced in the past, illustrating the need for further research into 3D adversarial attacks.

    %\item To ensure reproducibility and facilitate further research, we provide a publicly available implementation of \textit{AdvIRL} at \url{https://github.com/Tommy-Nguyen-cpu/AdvIRL/tree/MultiView-Clean}.
\end{enumerate}

\iffalse
\textbf{Contributions3}
\begin{enumerate}
\item We introduce \textit{AdvIRL}, the first black-box adversarial framework for Neural Radiance Fields (NeRF) that:
    \begin{itemize}
    \item Generates robust adversarial noise resilient to 3D transformations
    \item Enables high-confidence targeted misclassifications across diverse scenes
    \item Provides a novel approach for adversarial training of vision systems
    \end{itemize}

\item We demonstrate the framework's effectiveness by successfully attacking vision models across multiple domains, including:
    \begin{itemize}
    \item Small-scale objects (e.g., transforming a banana's classification)
    \item Large environmental scenes (e.g., manipulating lighthouse representations)
    \item Achieving misclassifications that highlight critical vulnerabilities in AI vision models
    \end{itemize}

\item We present a reproducible, open-source implementation that:
    \begin{itemize}
    \item Leverages Instant Neural Graphics Primitives for noise generation
    \item Utilizes reinforcement learning for adversarial model crafting
    \item Facilitates future research in 3D adversarial machine learning
    \end{itemize}
\end{enumerate}
\fi

%SECTION 2 proposal:
\section{Method}
In this section, we introduce our proposed framework, \textit{AdvIRL}, designed to generate adversarial noise for 3D models. Unlike traditional adversarial methods, our framework operates on a higher dimension leading to more robust results. We describe the components apart of our AdvIRL pipeline in the following subsections.
%Our proposed framework, AdvIRL, generates adversarial noise for 3D neural radiance field (NeRF) models through a three-step process. The key steps include using CLIP for image classification, leveraging Instant-NGP to render 3D objects from multiple perspectives, and applying reinforcement learning to manipulate parameters to produce adversarial examples. In this section, we detail each stage and the reinforcement learning (RL) specifics that enable effective black-box adversarial attacks.

\begin{figure*}[h]
    \centering
    \includegraphics[width=6.0in]{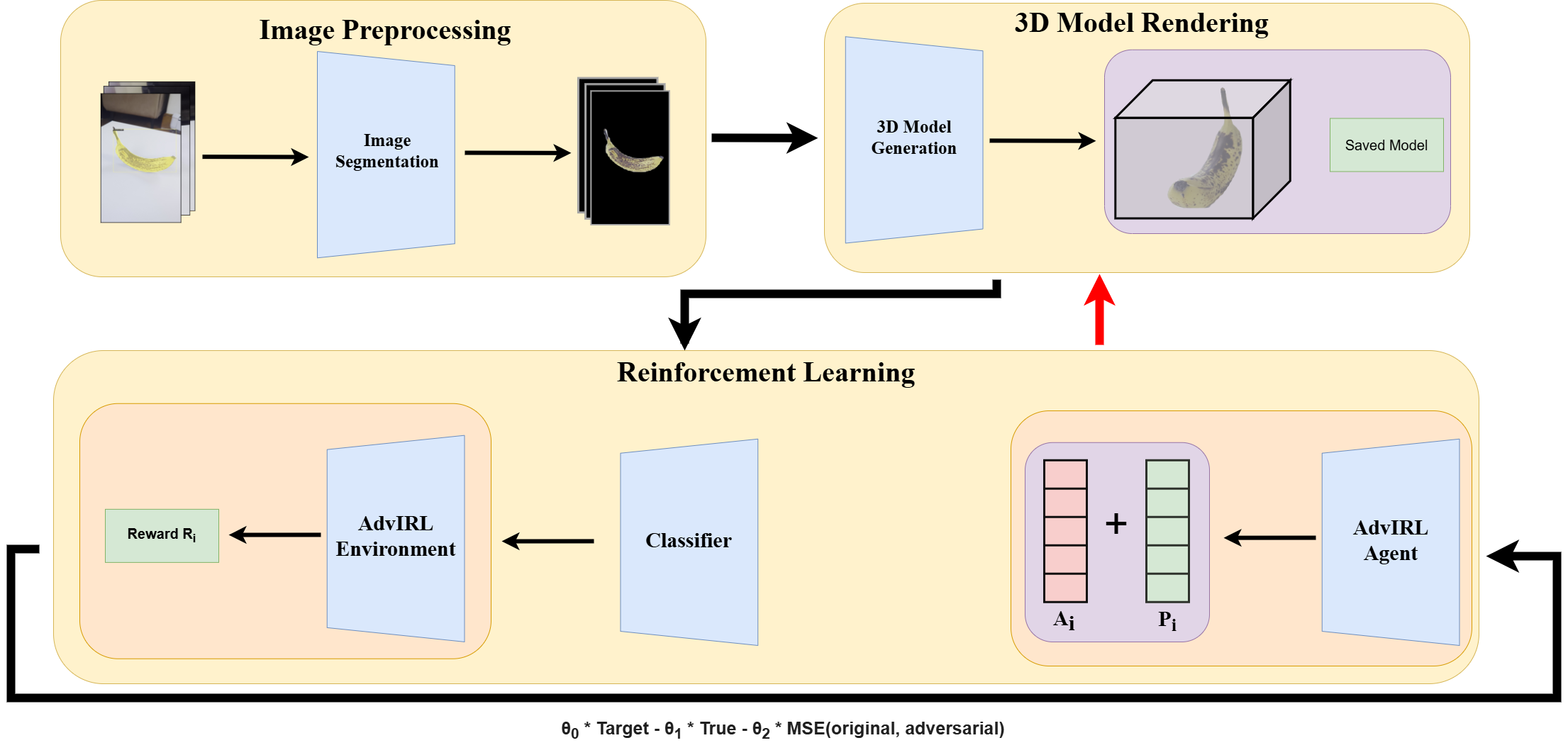}
    \caption{AdvIRL begins by processing a set of input images to generate segmented images, denoted as $X^{\text{segmented}}$. These segmented images are then used to render the NeRF model, producing rendered images $X$. The initial parameters $P_0$ of the NeRF model are extracted, enabling AdvIRL to modify them as part of the pipeline. Using the initial observation space ($X$), AdvIRL generates an action $A_i$, which is concatenated with the parameters $P_i$ at timestep $i$. The updated parameters are subsequently processed by Instant-NGP to generate multi-view shots of the object. These multi-view shots are then classified using our CLIP classifier, which outputs results that the environment uses to compute a reward $R$, as defined in the accompanying figure. This reward guides the agent in determining subsequent actions. The red arrow in the diagram illustrates the feedback loop, where parameters generated by AdvIRL are fed back to Instant-NGP to produce the adversarial 3D model.}
    %\caption{Given a saved instance of the NeRF model produced by Instant-NGP, the agent produces action $A_i$ which gets passed to the environment $E$ and concatenated with parameters $P_i$. The updated parameters are then fed through Instant-NGP which generates multi-view shots of our object. Next, these multi-view shots are fed to our CLIP classifier. The results of the classifier are subsequently used by the environment to produce reward $R$ as defined in the figure. The reward is used by the agent to determine future actions. The red arrow indicates parameters generated by AdvIRL being fed back to Instant-NGP to generate the adversarial 3D model.}
    \label{fig:ModifyNeRFParams}
\end{figure*}

The adversarial noise generation process comprises four primary steps. First, input images are processed through an image segmentation algorithm, generating segmented images $X$. Secondly, segmented images $X$ are classified using the Contrastive Language-Image Pretraining (CLIP) model to ensure only correctly classified images are used \citep{radford2021learning}. Next, Instant-NGP is employed to render the 3D object from various viewpoints, creating snapshots of the NeRF model. Finally, AdvIRL is configured with appropriate parameters to generate adversarial examples from different perspectives. Figure \ref{fig:ModifyNeRFParams} illustrates how AdvIRL adjusts the parameters of Instant-NGP to generate adversarial outcomes.

\iffalse
%ME:
\subsection{Adversarial Noise Generation Process}
To determine our noise generation process, we conducted an extensive hyperparameter exploration:

% \begin{itemize}
    % \item 
    \subsubsection{Reward Function Design:} We iteratively refined our reward function by testing multiple configurations, ultimately selecting weights that balance:
    \begin{enumerate}
        \item Misclassification confidence
        \item Minimal visual perturbation
        \item Consistent adversarial effectiveness across viewing angles
    \end{enumerate}
    
    % \item 
    \subsubsection{Hyperparameter Selection:} Our hyperparameters were chosen through grid search and Bayesian optimization, considering computational constraints and attack efficacy.
    
    % \item 
    \subsubsection{Constraint Reasoning:} The small n\_steps and batch\_size were not just hardware limitations but strategic choices to prevent gradient instability in high-dimensional parameter spaces.
% \end{itemize}
\fi

\subsection{Image Segmentation}
AdvIRL includes an image preprocessing step before entering the reinforcement learning cycle. As part of this process, an image segmentation algorithm, Detectron2  \citep{wu2019detectron2}, is utilized to generate a mask ($M$) and class predictions ($P$) based on the input images. These outputs are then used to create segmented images ($X$). By leveraging a pretrained Mask R-CNN model with a ResNet-50 backbone, we achieve strong results without requiring additional training.

\subsection{Adversarial Generation using Reinforcement Learning}
Once the segmented images $X$ are generated, they serve as the initial \textbf{observation space}. Based on the current state, AdvIRL generates an \textbf{action} $A$, which is concatenated with the existing Instant-NGP parameters $P^{\text{old}}$ to produce adversarial parameters $P^{\text{new}}$. Instant-NGP then uses $P^{\text{new}}$ to generate adversarial images $X^{\text{adv}}$ and the adversarial NeRF model $M^{\text{adv}}$. The adversarial images $X^{\text{adv}}$ are subsequently processed by our CLIP classifier, producing a nested list where each entry contains the label and its corresponding classification confidence. The nested list is then processed to produce a dictionary $D$ containing the label (\textbf{key}) and a list containing the average confidence and number of images predicted for this label (\textbf{value}).

Once preprocessed, dictionary $D$, input images $X$ and adversarial images $X^{\mathrm{adv}}$ will be used to generate reward $R$ formulated as:

\iffalse
AdvIRL utilizes reinforcement learning to generate adversarial noise that is resilient to diverse transformations. We detail the observation and action spaces, reward system, and parameters used for optimization below.

\textbf{Observation \& Action Space} To address the challenges of adversarial attacks in 3D space, the observation space consists of images of the scene captured from multiple viewpoints. AdvIRL analyzes these images to generate actions that modify Instant-NGP parameters. Our observation space is limited to 20 images per scene to comply with hardware constraints, as described in Section~\ref{sec:Constraints}.

\textbf{Reward System} The reward function encourages the generation of adversarial noise that is effective yet minimally disruptive to the original scene. We define the reward as:
\fi

% \begin{equation}
%     \begin{split}
%         reward(True, Target, X, X^{\text{adv}}) = \\ Target * \theta_0 - True * \theta_1 - MSE(X, X^{\text{adv}}) * \theta_2
%     \end{split}
%     \label{equation_1}
% \end{equation}
\begin{align}
    \mathrm{reward}&(\mathrm{True}, \mathrm{Target}, X, X^{\mathrm{adv}}) = \\
    &\quad \mathrm{Target} \cdot \theta_0 - \mathrm{True} \cdot \theta_1 \nonumber \\
    &\quad - \mathrm{MSE}(X, X^{\mathrm{adv}}) \cdot \theta_2
    \label{eq:equation_1}
\end{align}

Here, $\theta_0$ and $\theta_1$ are hyperparameters that balance the confidence of the true predictions of the target class and the true predictions. The term $MSE(X, X^{\text{adv}})$ penalizes excessive modifications, encouraging AdvIRL to focus noise generation within the object's bounds. In our experiments, we set $\theta_0 = 100$, $\theta_1 = -1$, and $\theta_2 = 0.00005$. This trade-off prevents extreme image alterations while ensuring effective adversarial noise generation, high misclassification confidence, and robust adversarial results across vantage points.

The adversarial images $X^{\text{adv}}$ and the reward $R$ are utilized by AdvIRL to determine subsequent actions. Algorithm \ref{alg:advirl} outlines the processes executed within a single timestep.

It is worth emphasizing that, unlike traditional adversarial methods that heavily depend on target model gradients, AdvIRL operates without using gradients at all. It also does not assume knowledge of the target model's architecture, sidestepping the limitations of current approaches that require gradient access or depend on transferability for black-box attacks. By incorporating reinforcement learning and leveraging agent exploration, AdvIRL relies solely on input images and classification labels to assess whether its actions move closer to achieving its adversarial objective. By relying exclusively on input images and classification labels, AdvIRL can be seamlessly transferred and applied to any vision model without requiring prior training on a separate model.

\begin{algorithm}
% \caption{AdvIRL} 
\caption{The AdvIRL Framework: Key steps for generating adversarial perturbations in neural radiance fields.}
\label{alg:advirl}
\textbf{Input:} Action $A$ \\
\textbf{Parameters:} Original images $X$, number of images $N$, target class confidence weight $\theta_0$, true class confidence weight $\theta_1$, MSE weight $\theta_2$ \\
\textbf{Output:} Adversarial images $X^{\mathrm{adv}}$, reward $R$ 

\begin{algorithmic}[1]
\STATE Update Instant-NGP parameters: 
\[
P^{\mathrm{new}} \gets A + P^{\mathrm{old}}
\]
\STATE Generate adversarial images: 
\[
X^{\mathrm{adv}} \gets \mathrm{NGP}(P^{\mathrm{new}})
\]

\STATE Obtain class predictions for $X^{\mathrm{adv}}$: 
\[
\mathrm{Predictions} \gets \mathrm{CLIP}(X^{\mathrm{adv}})
\]

\STATE Initialize class confidence dictionary: $D \gets \{\}$
\FOR{$n = 0$ to $N-1$}
    \STATE Extract predicted class:
    \[
    \mathrm{Class} \gets \mathrm{Predictions}(n)[1]
    \]
    \STATE Extract confidence score: 
    \[
    \mathrm{Conf} \gets \mathrm{Predictions}(n)[2]
    \]
    \IF{$\mathrm{Class}$ not in $D$}
        \STATE Initialize dictionary entry: 
        \[
        D[\mathrm{Class}] \gets [\mathrm{Conf}, 1]
        \]
    \ELSE
        \STATE Compute updated average confidence: 
        \[
        \mathrm{Conf}^{\mathrm{sum}} \gets D[\mathrm{Class}][0] \cdot D[\mathrm{Class}][1] + \mathrm{Conf}
        \]
        \STATE Update count for $\mathrm{Class}$: 
        \[
        D[\mathrm{Class}][1] \gets D[\mathrm{Class}][1] + 1
        \]
        \STATE Calculate new average confidence:
        \[
        \mathrm{Conf}^{\mathrm{avg}}_{\mathrm{new}} \gets \frac{\mathrm{Conf}^{\mathrm{sum}}}{D[\mathrm{Class}][1]}
        \]
        \STATE Update dictionary: 
        \[
        D[\mathrm{Class}] \gets [\mathrm{Conf}^{\mathrm{avg}}_{\mathrm{new}}, D[\mathrm{Class}][1]]
        \]
    \ENDIF
\ENDFOR

\STATE Compute reward $R$: 
\[
R = \mathrm{Target} \cdot \theta_{0} - \mathrm{True} \cdot \theta_{1} - \mathrm{MSE}(X, X^{\mathrm{adv}}) \cdot \theta_{2}
\]
\STATE Return $X^{\mathrm{adv}}, R$

\end{algorithmic}
\end{algorithm}

\subsection{Experimental Setup}
We target the CLIP Resnet50 model \cite{dosovitskiy2021} due to its flexible class labels, allowing us to define any set of string labels for classification. This flexibility is crucial for generating adversarial examples across varied scenarios.

Our experiments utilize the Tanks and Temples dataset \citep{Knapitsch2017}, as well as an additional banana scene collected by our team. The Tanks and Temples dataset offers diverse 3D scenes, enabling a thorough evaluation of AdvIRL's performance across different settings.

\subsubsection{Hyperparameter Selection:} Our hyperparameters were chosen through grid search and Bayesian optimization, considering computational constraints and attack efficacy.

\subsubsection{Policy Gradient Algorithm} We selected Proximal Policy Optimization (PPO) as our policy gradient algorithm for its simplicity and robustness. The following parameters were used for our experiments:
\begin{itemize}
    \item \textbf{n\_steps}: Set to 2.
    \item \textbf{batch\_size}: Set to 2.
    \item \textbf{max\_grad\_norm}: Set to 0.00001.
\end{itemize}
Our method’s reliance on small n\_steps, batch\_size, and max\_grad\_norm were not just hardware limitations caused by the large action space and the memory-intensive nature of rendering NeRF model, but strategic choices to prevent gradient instability in high-dimensional parameter spaces.

\subsection{Constraints}\label{sec:Constraints}
\subsubsection{Hardware}: AdvIRL was trained on a workstation with two GPUs, each with 24 GB of VRAM. Instant-NGP and the CLIP classifier operated on separate GPUs, while the reinforcement learning model ran on the CPU.

\subsubsection{Time Complexity}: Due to the iterative nature of NeRF rendering at each timestep, training AdvIRL took an average of two days for targeted attacks. Untargeted attacks, being less constrained, required less training time.

\section{Results}
%ME: will split into scenes for ease of reading

\begin{figure*}
    \centering
    \begin{subfigure}[b]{0.165\textwidth}
        \includegraphics[width=\textwidth]{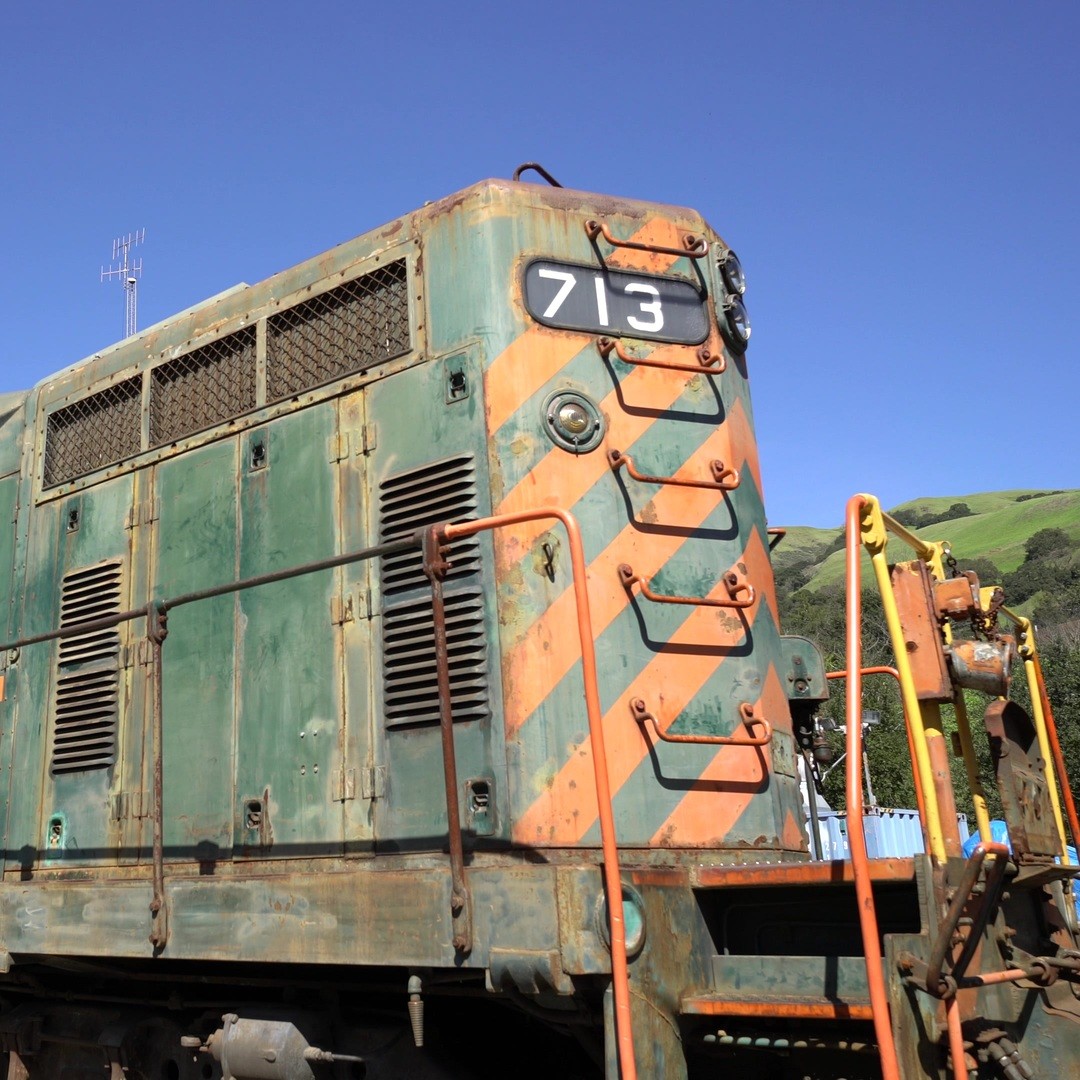}
        \label{fig:train_og}
    \end{subfigure}
    \begin{subfigure}[b]{0.165\textwidth}
        \includegraphics[width=\textwidth]{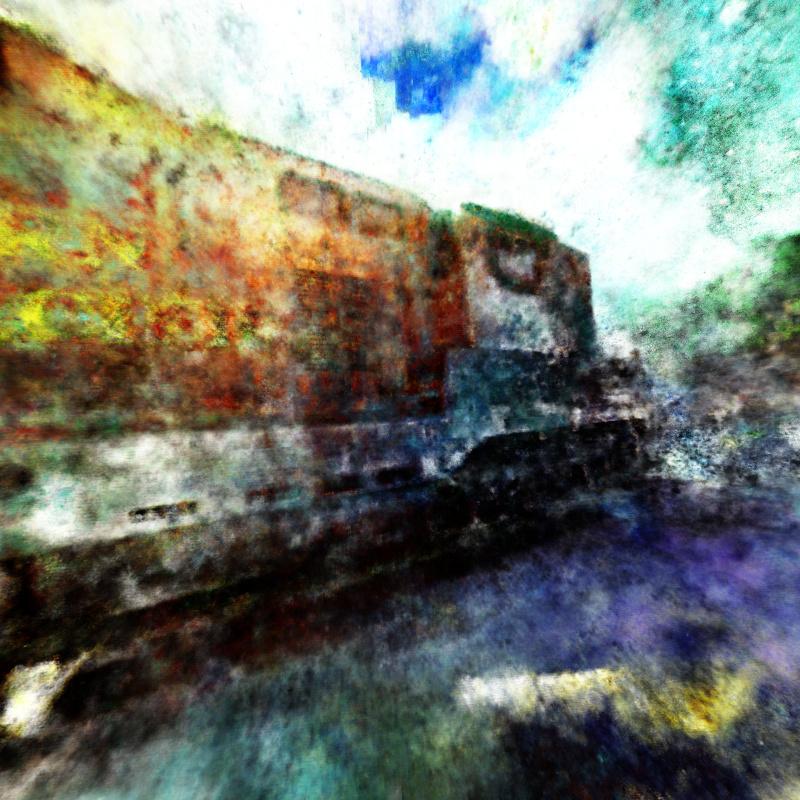}
        \label{fig:train_1}
    \end{subfigure}
    \begin{subfigure}[b]{0.165\textwidth}
        \includegraphics[width=\textwidth]{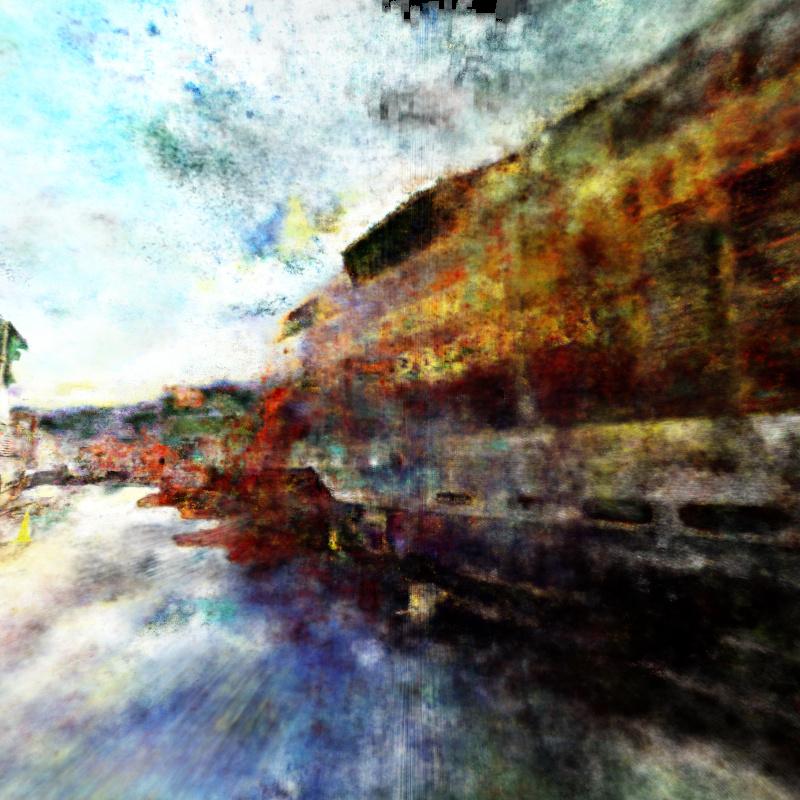}
        \label{fig:train_2}
    \end{subfigure}
    \begin{subfigure}[b]{0.165\textwidth}
        \includegraphics[width=\textwidth]{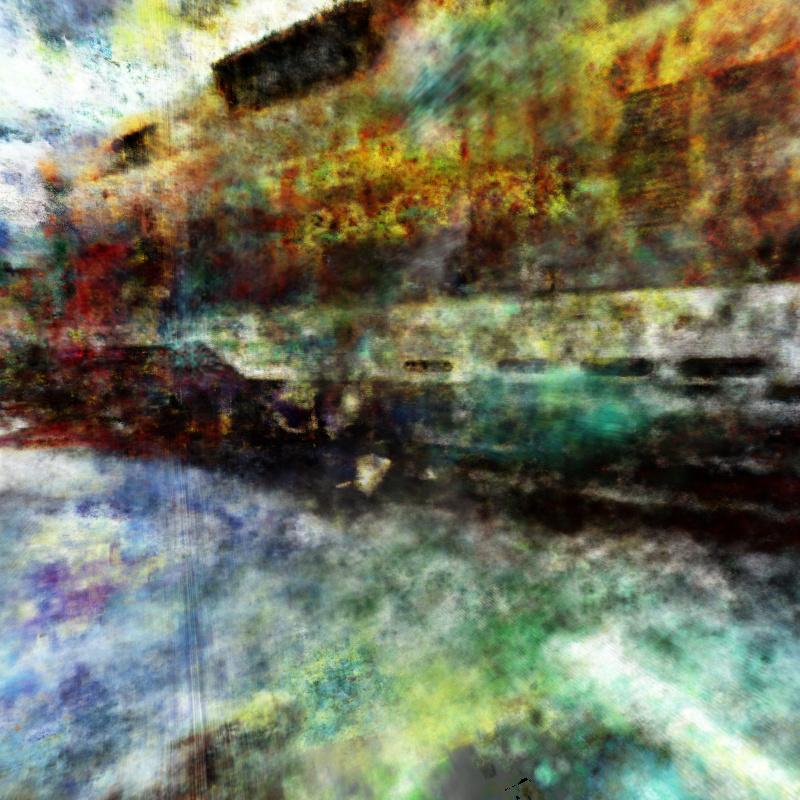}
        \label{fig:train_3}
    \end{subfigure}
    \begin{subfigure}[b]{0.165\textwidth}
        \includegraphics[width=\textwidth]{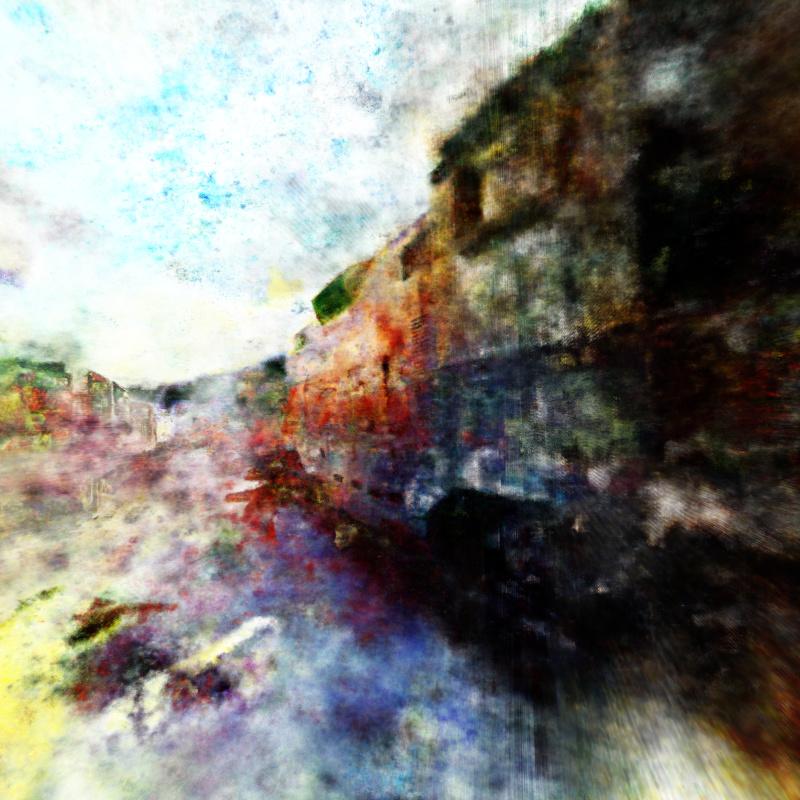}
        \label{fig:train_4}
    \end{subfigure}
    \begin{subfigure}[b]{0.165\textwidth}
        \includegraphics[width=\textwidth]{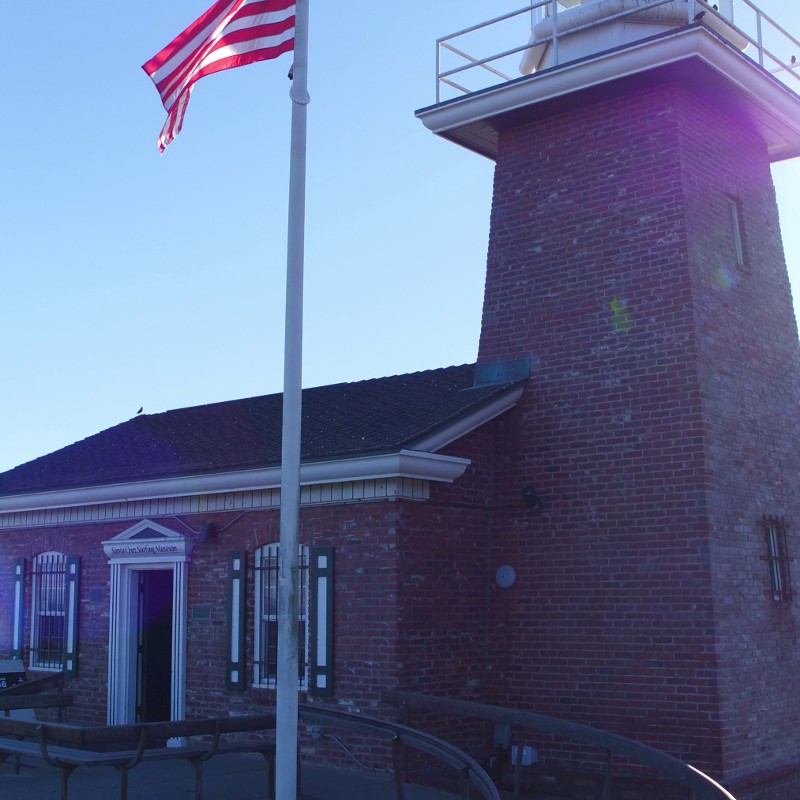}
        \label{fig:lighthouse_og}
    \end{subfigure}
 \begin{subfigure}[b]{0.165\textwidth}
        \includegraphics[width=\textwidth]{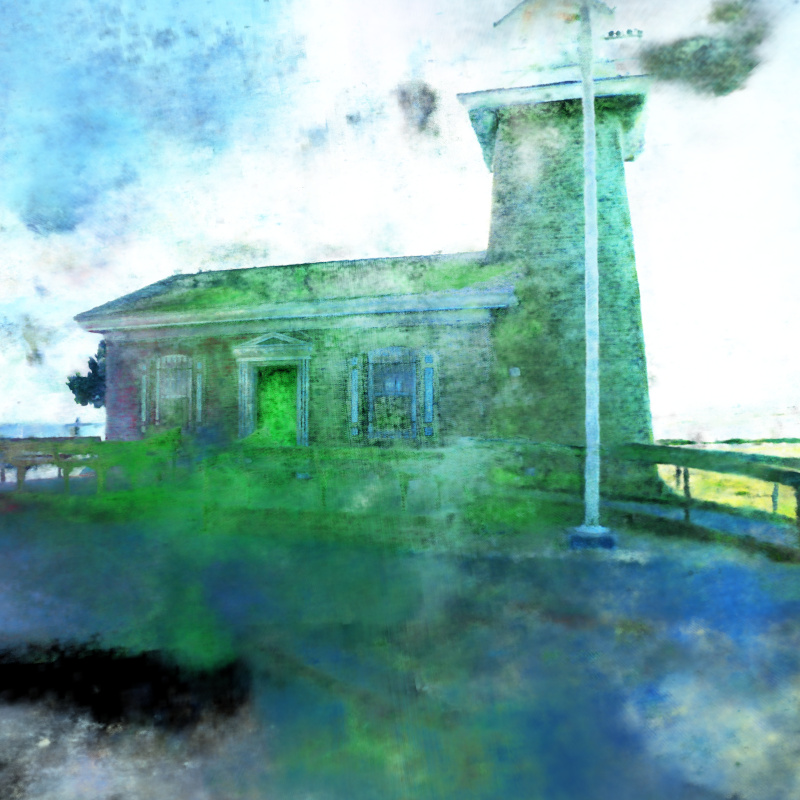}
        \label{fig:truckside}
    \end{subfigure}
    \begin{subfigure}[b]{0.165\textwidth}
        \includegraphics[width=\textwidth]{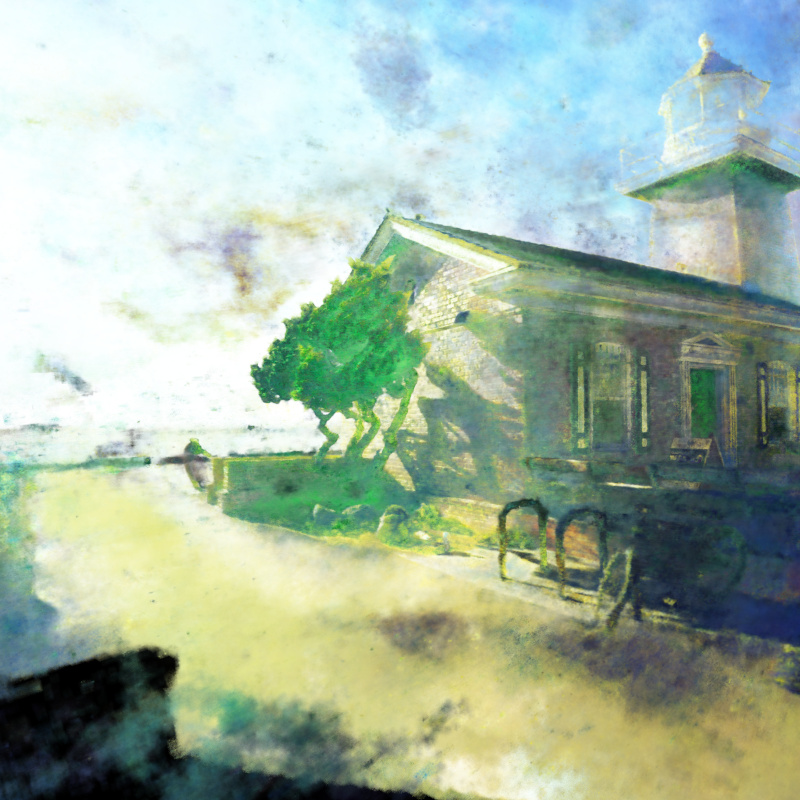}
        \label{fig:truckside2}
    \end{subfigure}
    \begin{subfigure}[b]{0.165\textwidth}
        \includegraphics[width=\textwidth]{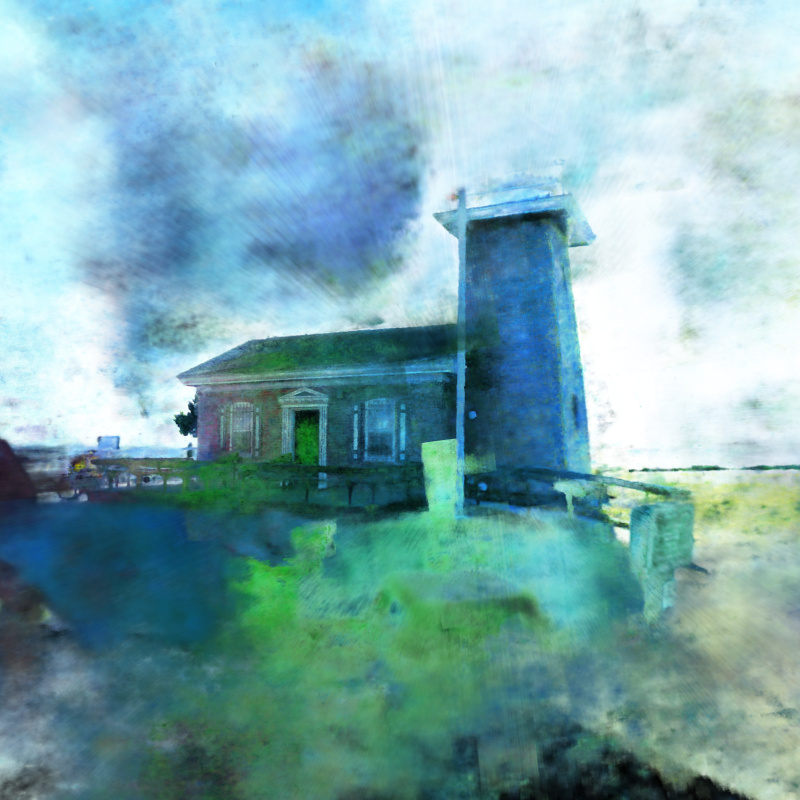}
        \label{fig:truckfront}
    \end{subfigure}
    \begin{subfigure}[b]{0.165\textwidth}
        \includegraphics[width=\textwidth]{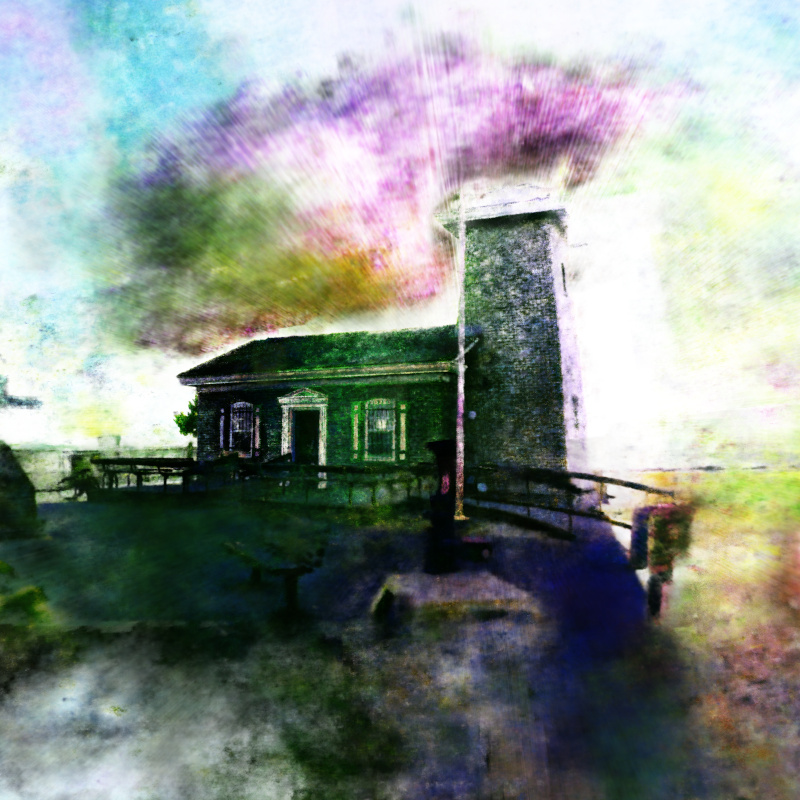}
        \label{fig:birdhouse8}
    \end{subfigure}
    \begin{subfigure}[b]{0.164\textwidth}
        \includegraphics[width=\textwidth]{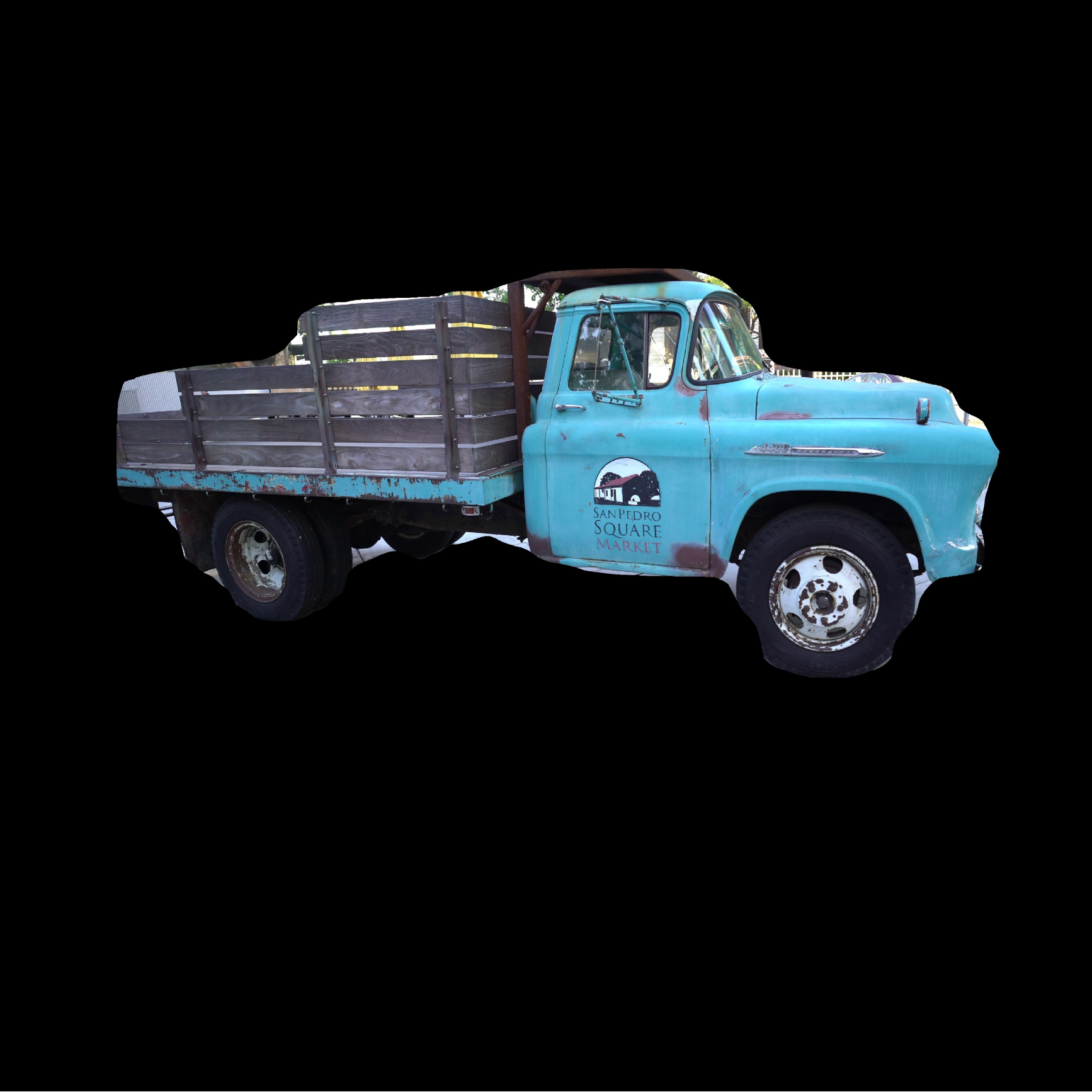}
        \label{fig:truck_og}
    \end{subfigure}
    \begin{subfigure}[b]{0.165\textwidth}
        \includegraphics[width=\textwidth]{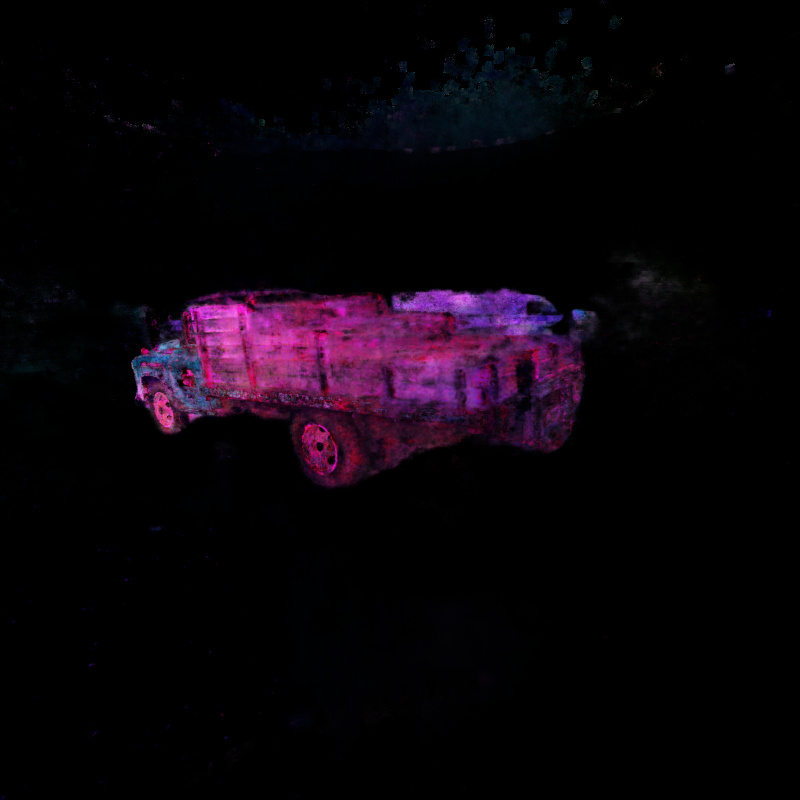}
        \label{fig:truckside3}
    \end{subfigure}
    \begin{subfigure}[b]{0.165\textwidth}
        \includegraphics[width=\textwidth]{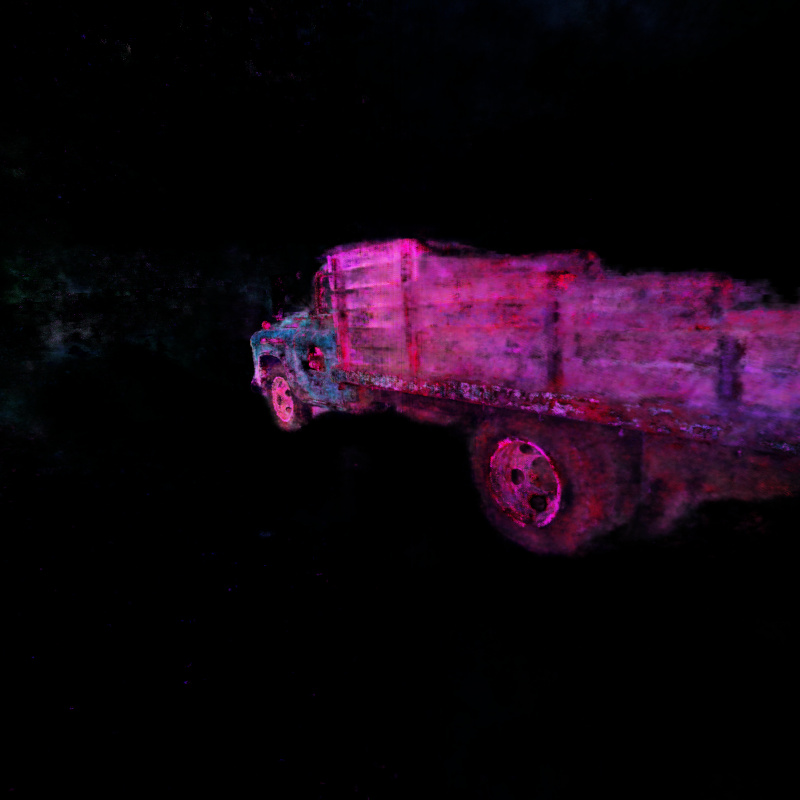}
        \label{fig:truckside4}
    \end{subfigure}
    \begin{subfigure}[b]{0.165\textwidth}
        \includegraphics[width=\textwidth]{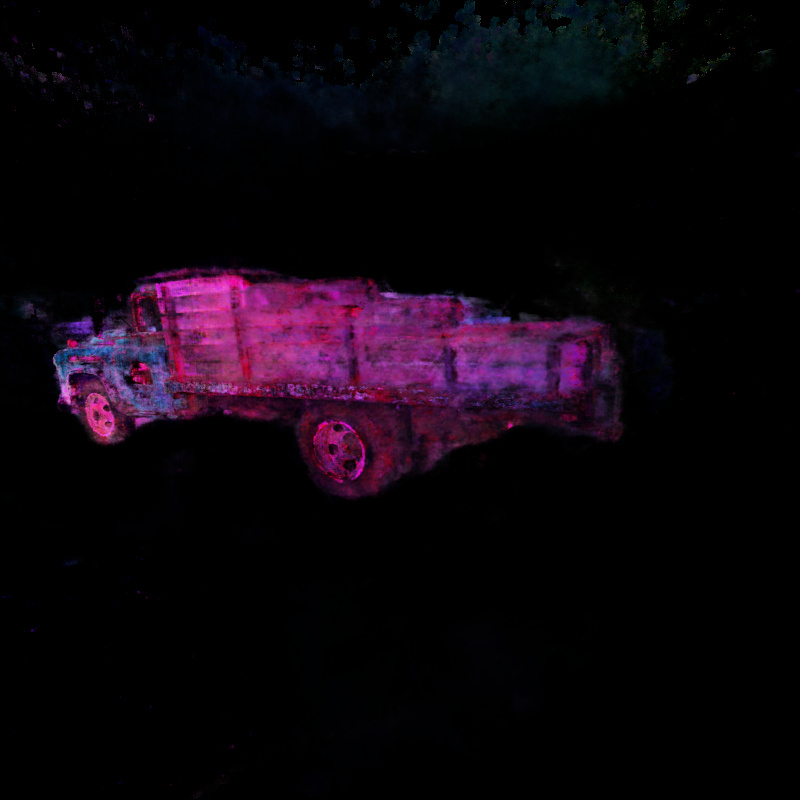}
        \label{fig:truckfront2}
    \end{subfigure}
    \begin{subfigure}[b]{0.165\textwidth}
        \includegraphics[width=\textwidth]{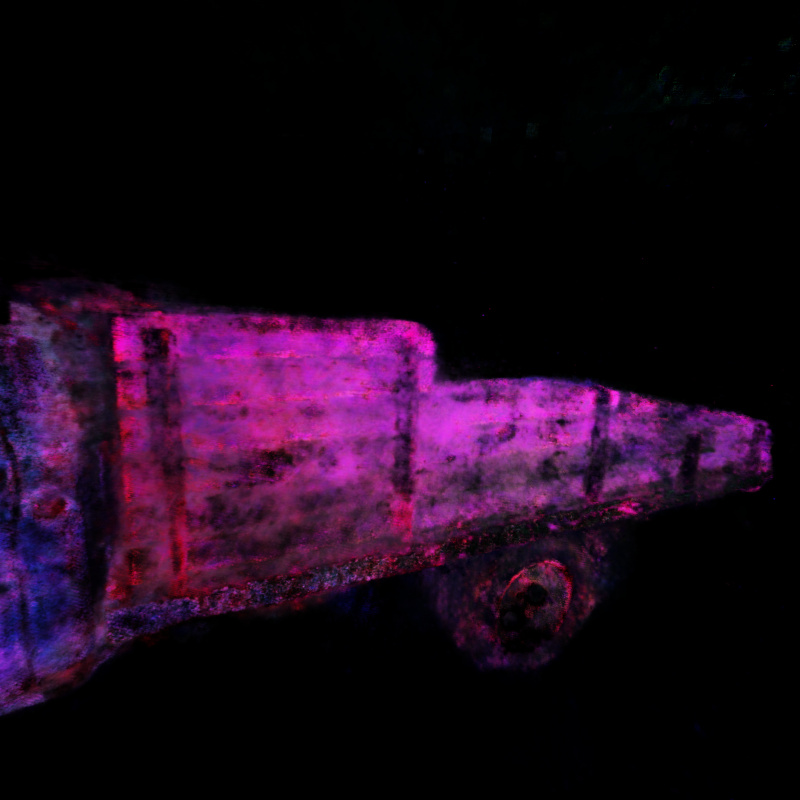}
        \label{fig:truckclose}
    \end{subfigure}
    
    \begin{subfigure}[b]{0.1578\linewidth}
        \includegraphics[width=\textwidth]{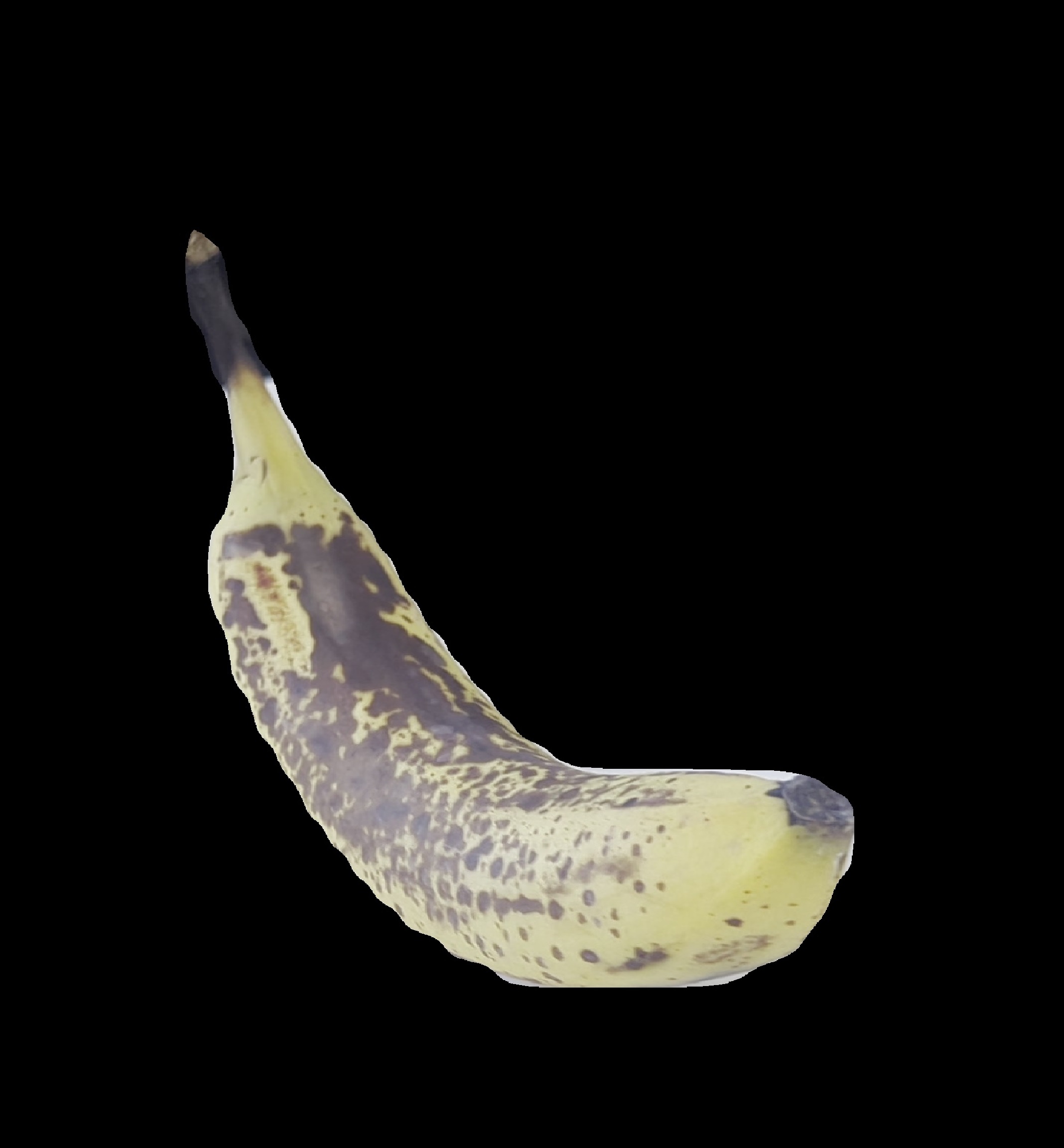}
        \label{fig:segmented_banana}
    \end{subfigure}
    \begin{subfigure}[b]{0.168\linewidth}
        \includegraphics[width=\textwidth]{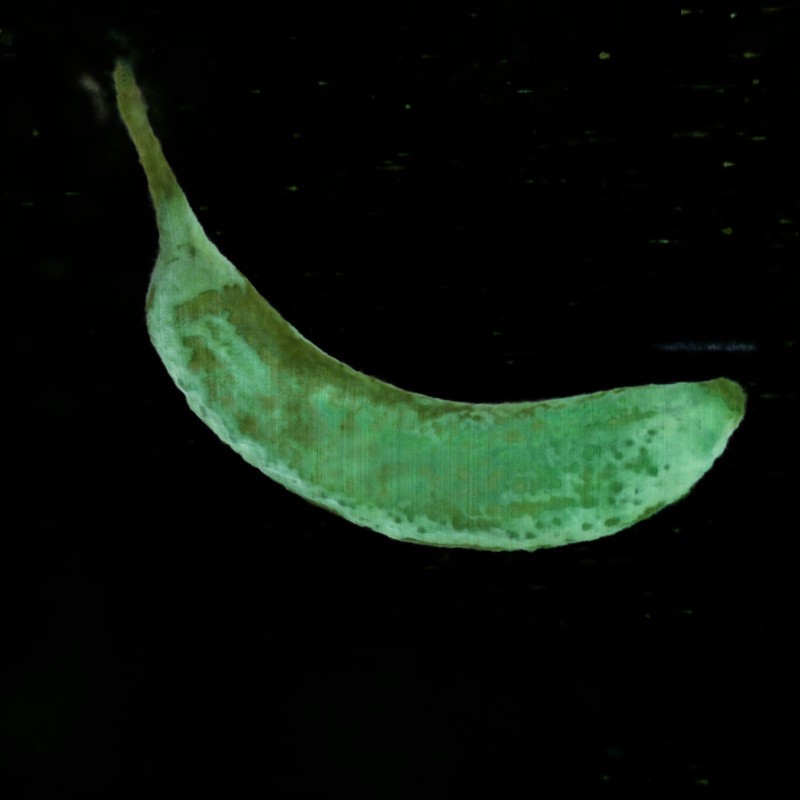}
        \label{fig:multi_angle_1}
    \end{subfigure}
    \begin{subfigure}[b]{0.168\linewidth}
        \includegraphics[width=\textwidth]{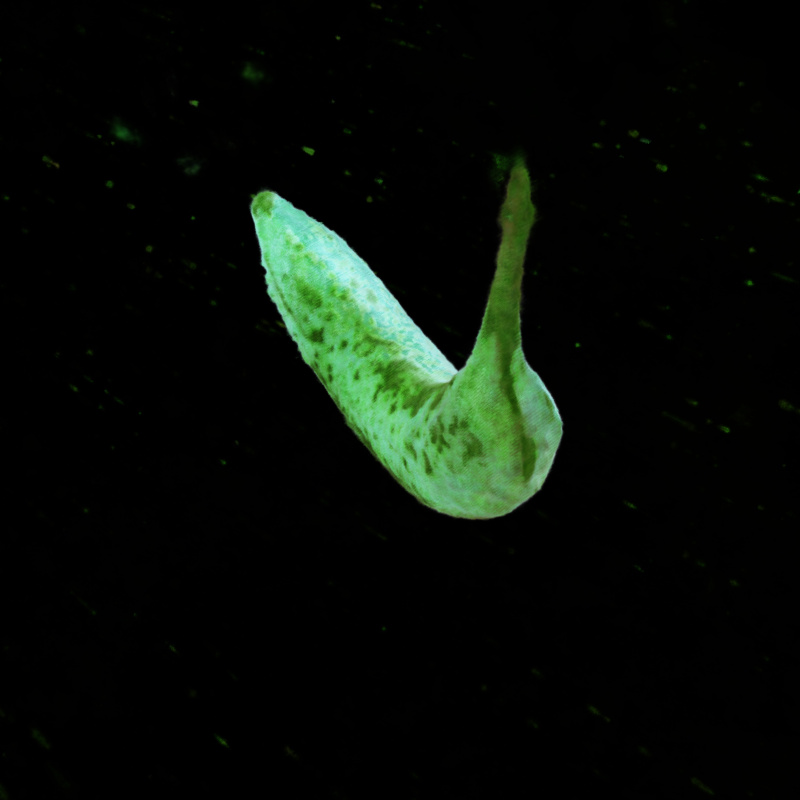}
        \label{fig:multi_angle_2}
    \end{subfigure}
    \begin{subfigure}[b]{0.168\linewidth}
        \includegraphics[width=\textwidth]{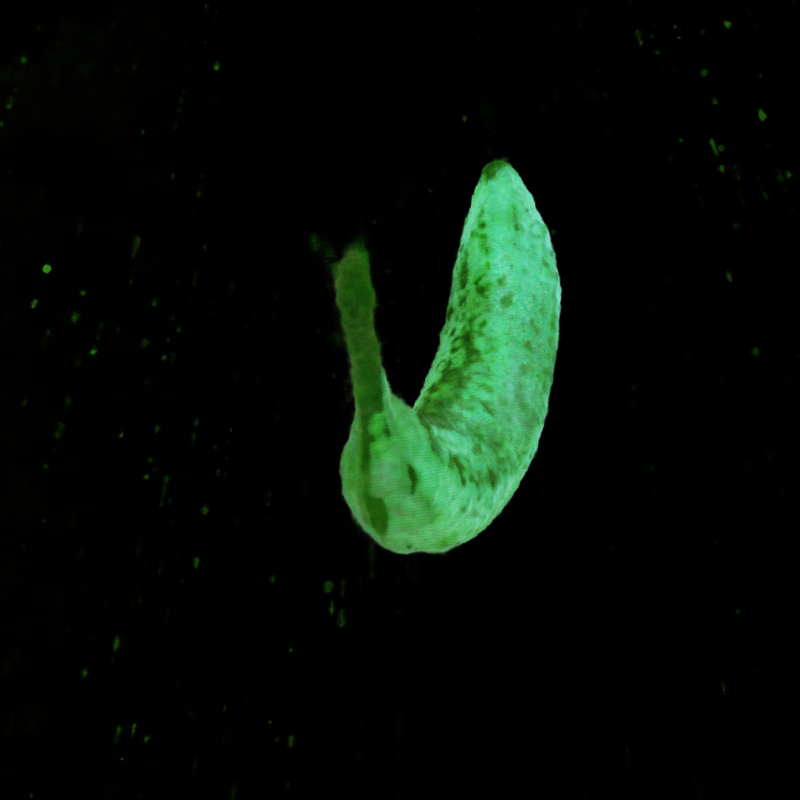}
        \label{fig:multi_angle_3}
    \end{subfigure}
    \begin{subfigure}[b]{0.168\linewidth}
        \includegraphics[width=\textwidth]{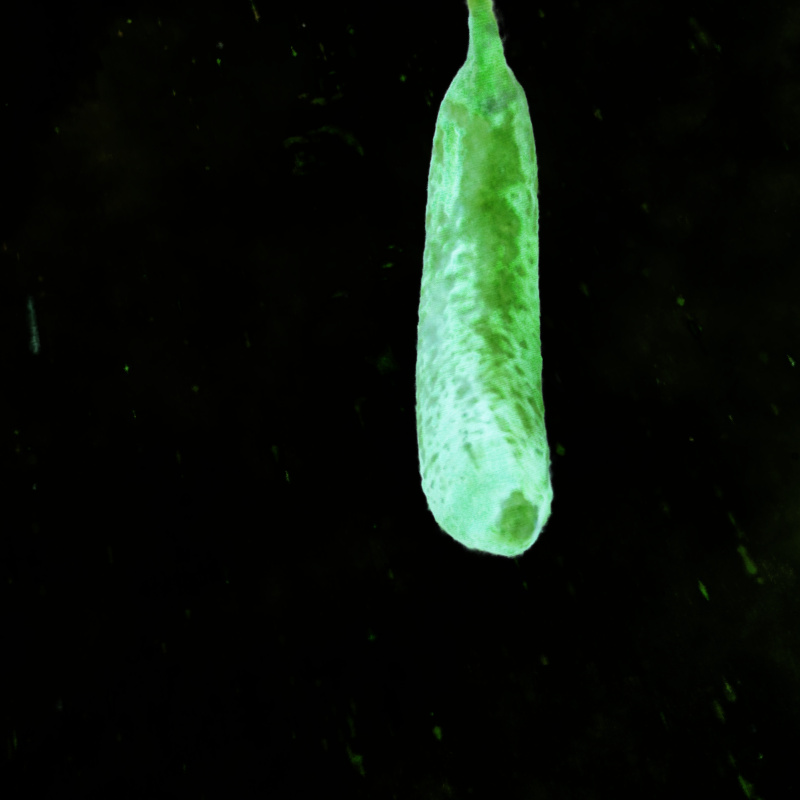}
        \label{fig:multi_angle_4}
    \end{subfigure}

    \begin{subfigure}[b]{0.165\linewidth}
        \includegraphics[width=\textwidth]{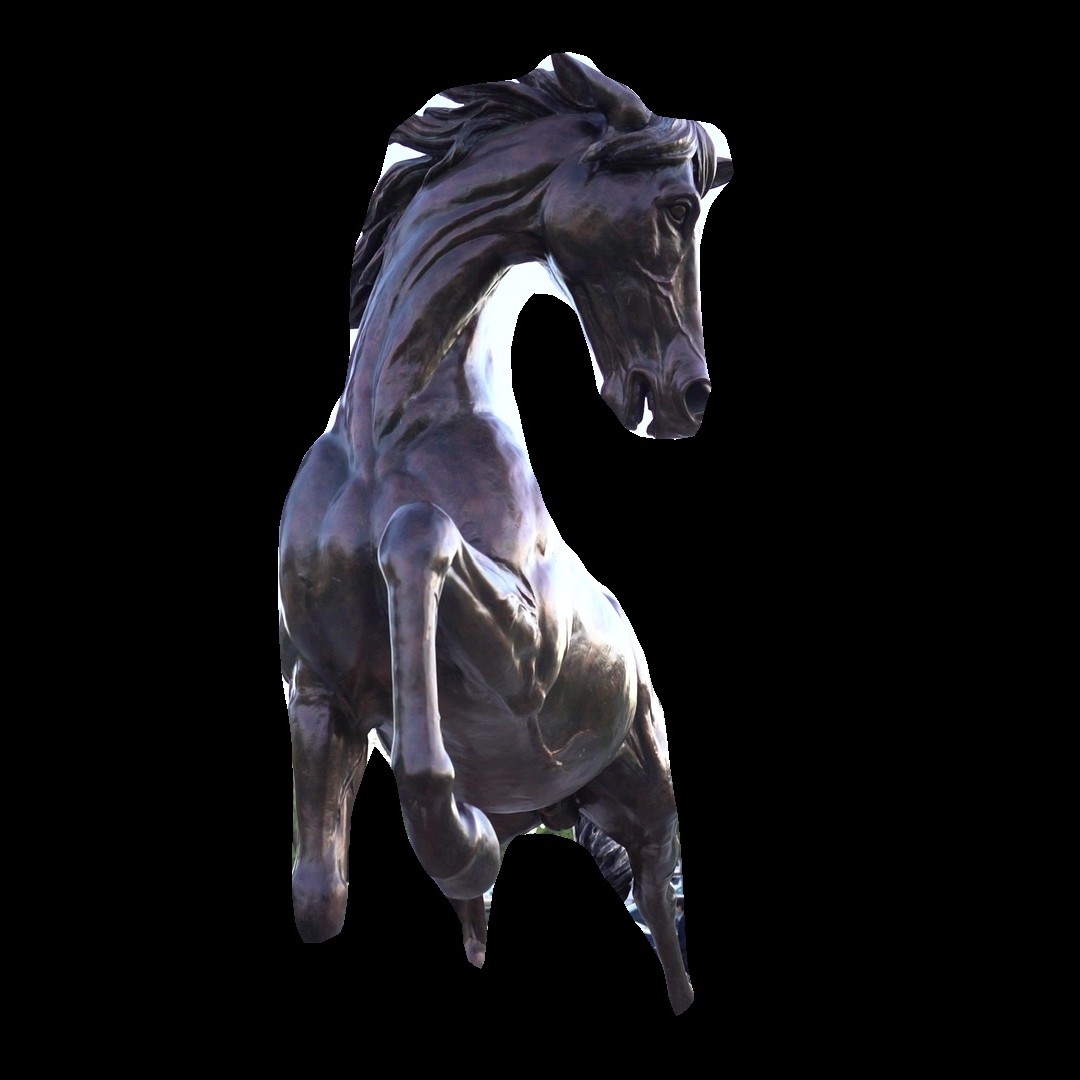}
        \label{fig:og_horse}
    \end{subfigure}
    \begin{subfigure}[b]{0.165\linewidth}
        \includegraphics[width=\textwidth]{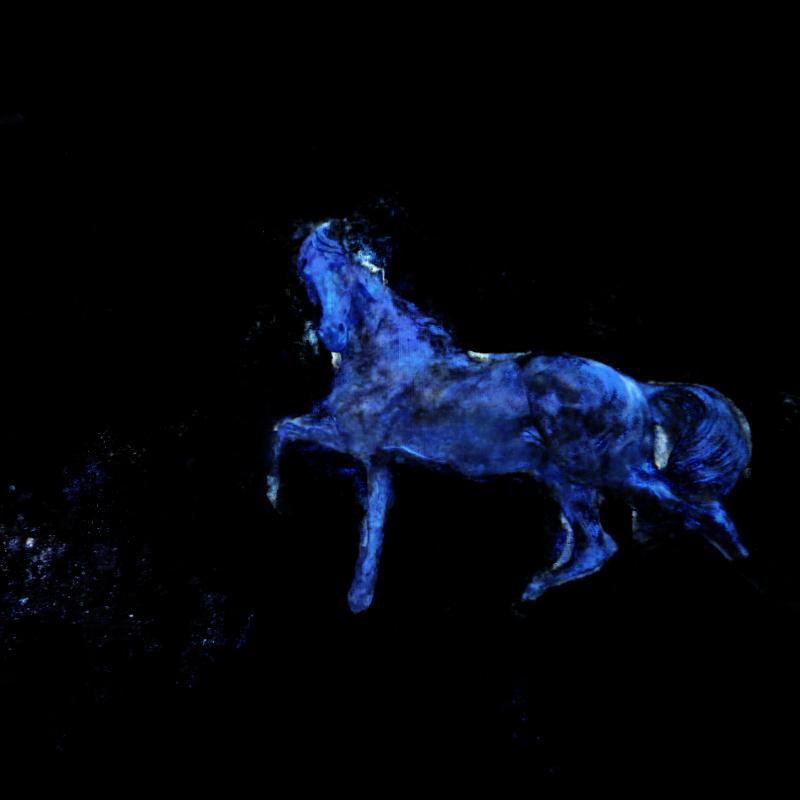}
        \label{fig:multi_horse_1}
    \end{subfigure}
    \begin{subfigure}[b]{0.165\linewidth}
        \includegraphics[width=\textwidth]{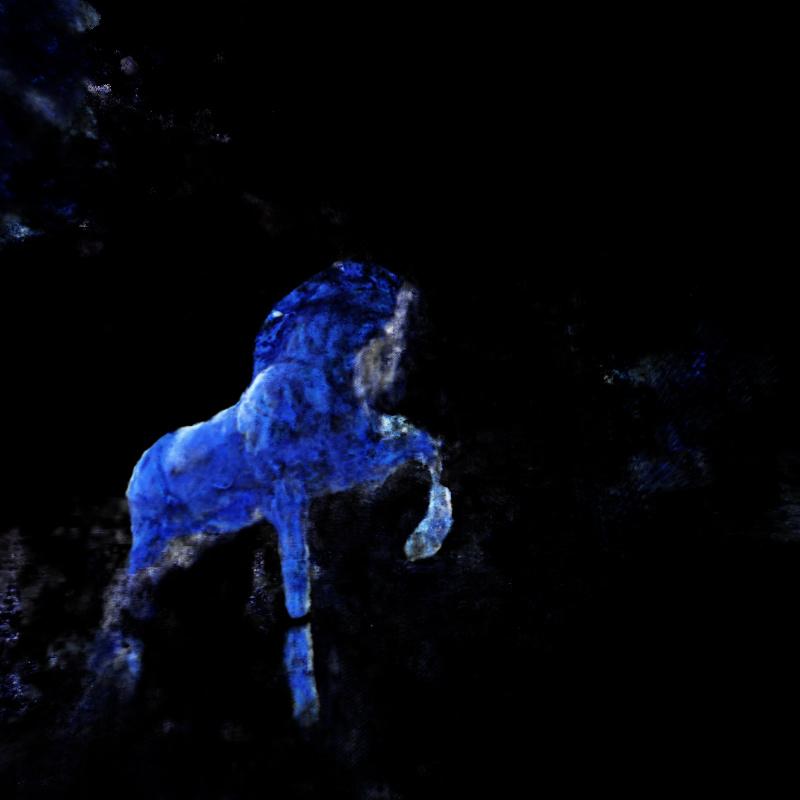}
        \label{fig:multi_horse_angle_2}
    \end{subfigure}
    \begin{subfigure}[b]{0.165\linewidth}
        \includegraphics[width=\textwidth]{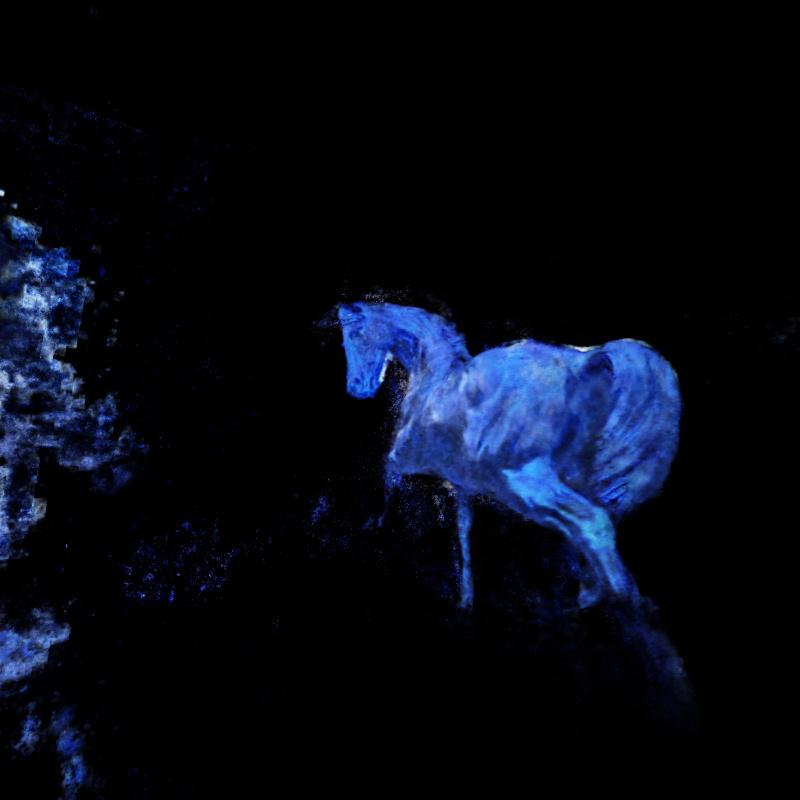}
        \label{fig:multi_horse_angle_3}
    \end{subfigure}
    \begin{subfigure}[b]{0.165\linewidth}
        \includegraphics[width=\textwidth]{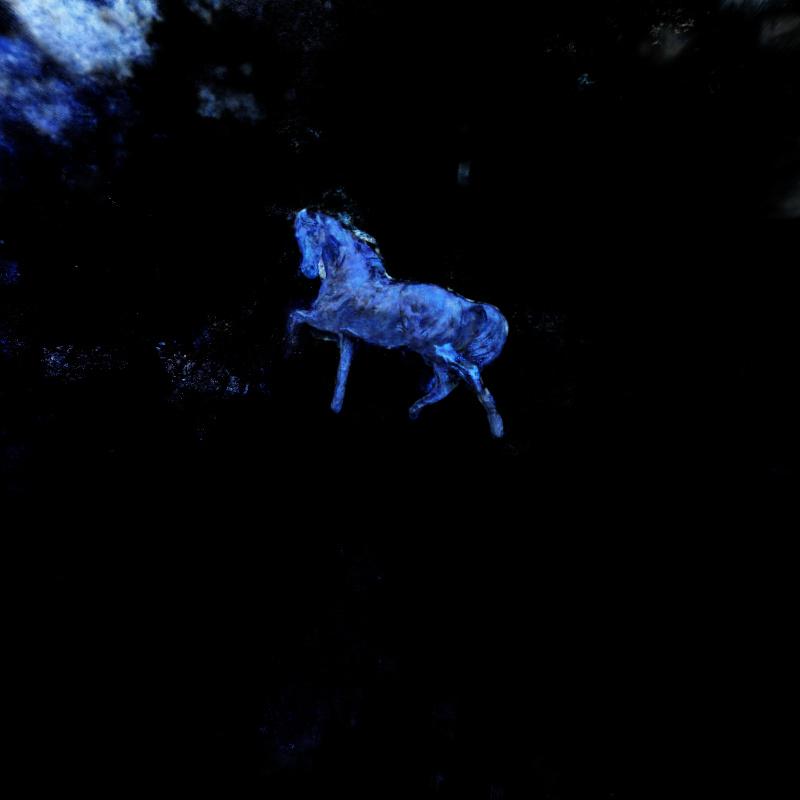}
        \label{fig:multi_horse_angle_4}
    \end{subfigure}
    \caption{Generated results of different scenes using AdvIRL, shown from multiple angles and distances. The leftmost column shows the original images, while subsequent columns show images with applied adversarial noise.}
    \label{fig:AllExperiments}
\end{figure*}

Our experiments evaluated AdvIRL's ability to generate adversarial noise for various scenes from the Tanks and Temples dataset \cite{Knapitsch2017}. Figure \ref{fig:AllExperiments} showcases the original scenes alongside the adversarially perturbed images produced by AdvIRL from different angles and distances.

\subsection{Adversarial Results without Segmentation}
Our initial experiments excluded image segmentation from our pipeline to assess the impact of AdvIRL on scenes with unrestricted freedom to modify their entirety. These tests were conducted on two scenes: the lighthouse and the train scene \cite{Knapitsch2017}.

\subsubsection{Lighthouse:} We first ran AdvIRL to generate untargeted adversarial noise. The average confidence across 13 images was 23.4\% for the boathouse class. Of the remaining images, six were classified as a lighthouse (19.4\% confidence), and one was misclassified as a church (18.2\% confidence). Additionally, targeted adversarial noise generation was applied with the boathouse class as the target.

\begin{figure}[h]
    \centering
    \includegraphics[width=.8\linewidth]{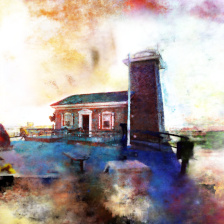}
    \caption{AdvIRL-generated adversarial noise targeting the boathouse class for the lighthouse scene, with 50\% classification confidence using the CLIP model.}
    \label{fig:Boathouse50percent}
\end{figure}

Figure \ref{fig:Boathouse50percent} shows the resulting adversarial image classified as a boathouse with 50\% confidence. Notably, the average confidence across all images was 20\%, indicating significant variability in classification confidence across different angles. Additional images of this experiment are shown in Figure \ref{fig:AllExperiments}. This result highlights AdvIRL's capability in generating class-targeted adversarial noise.

\subsubsection{Train Scene:} All 20 images were misclassified as various other objects (breakwater, megalith, couch, etc.) with classification confidences ranging from 4\% to 37\%. Unlike other scenes, the adversarial noise applied to the train scene significantly altered the object's identifying features, such as edges and text, which may have contributed to the high degree of misclassification.

\begin{figure}[h]
    \centering
    \begin{subfigure}[b]{0.235\linewidth}
        \includegraphics[width=\textwidth]{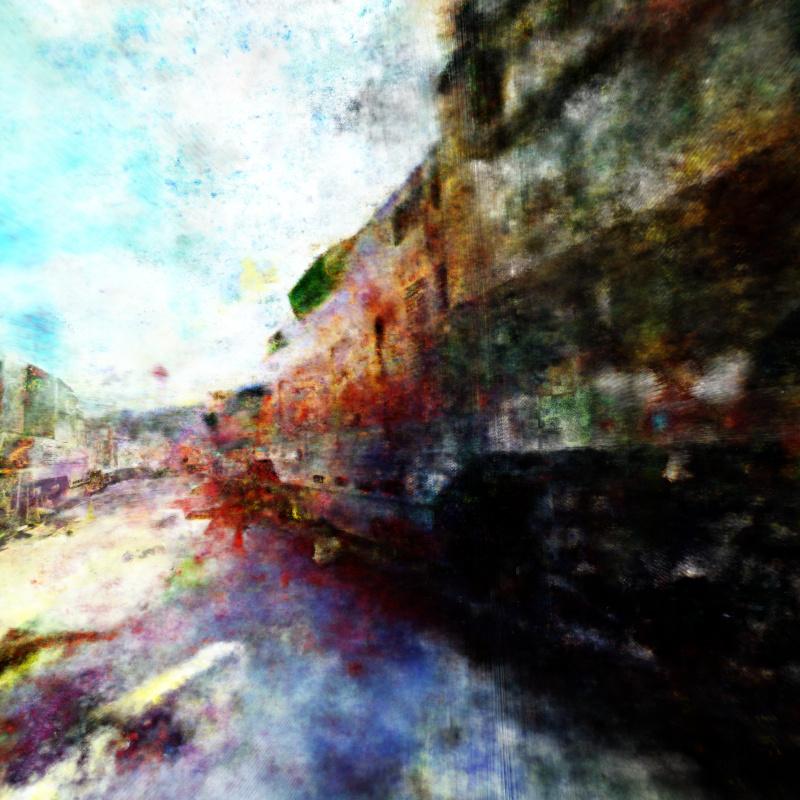}
        \caption{}
        \label{fig:train_multi_angle_1}
    \end{subfigure}
    \begin{subfigure}[b]{0.235\linewidth}
        \includegraphics[width=\textwidth]{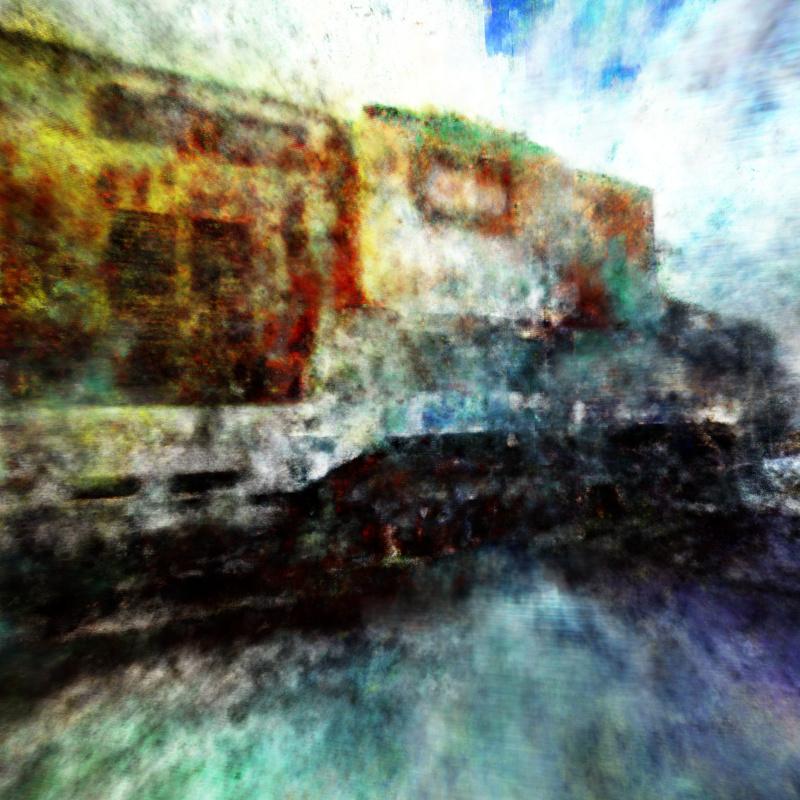}
        \caption{}
        \label{fig:train_multi_angle_2}
    \end{subfigure}
    \begin{subfigure}[b]{0.235\linewidth}
        \includegraphics[width=\textwidth]{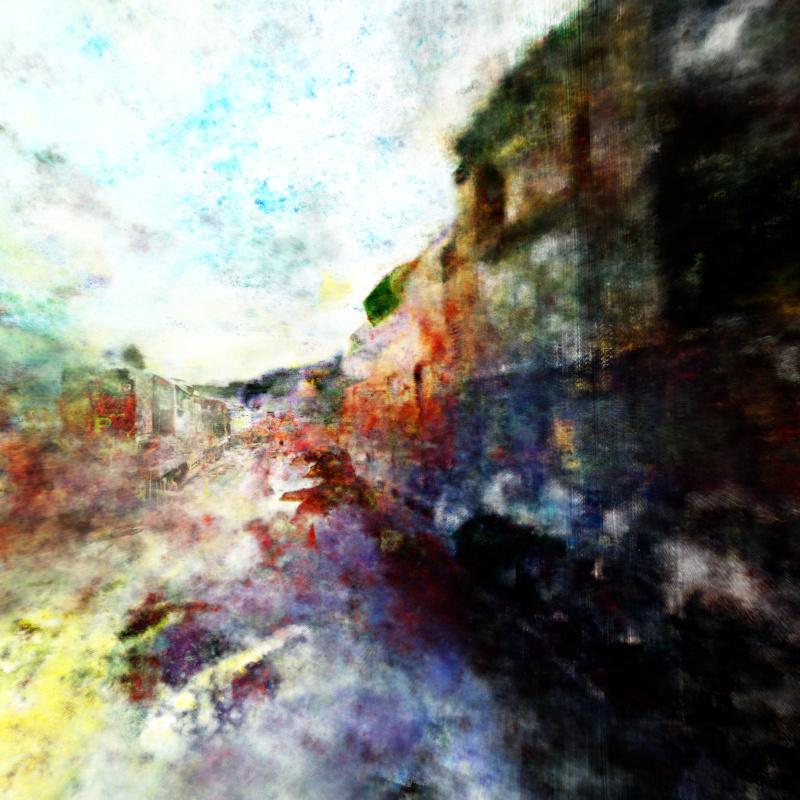}
        \caption{}
        \label{fig:train_multi_angle_3}
    \end{subfigure}
    \begin{subfigure}[b]{0.235\linewidth}
        \includegraphics[width=\textwidth]{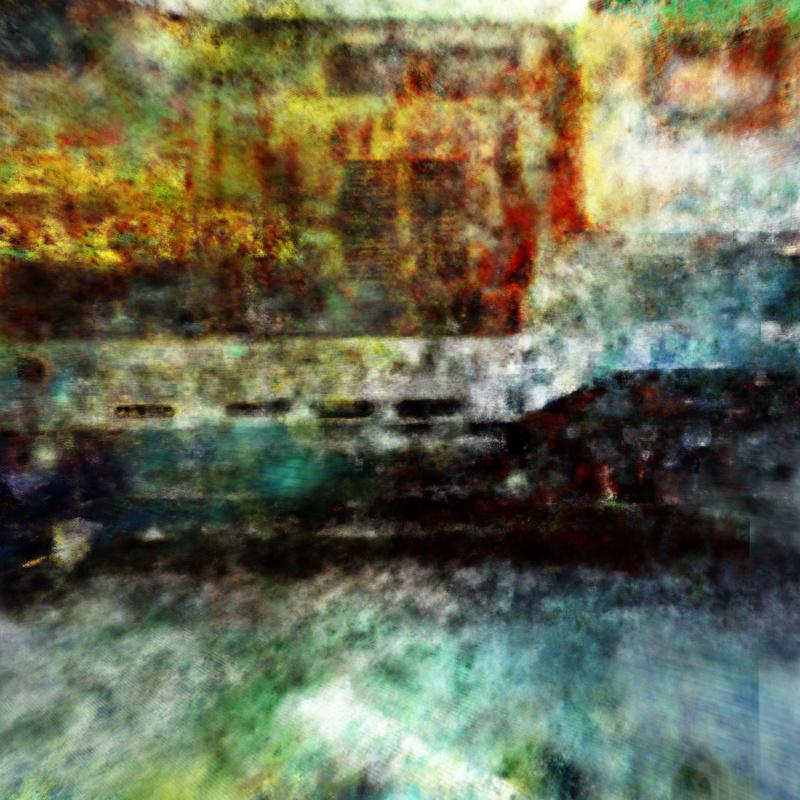}
        \caption{}
        \label{fig:train_multi_angle_4}
    \end{subfigure}
    \caption{Additional images of the adversarial train scene from various angles and distances.}
    \label{fig:Multi_Angle_train}
\end{figure}

\subsection{Adversarial Results For Segmented Input Images}
To prevent substantial alteration of the entire scene, our Segmentation-Reinforcement Learning pipeline guides AdvIRL in selectively modifying only the target object. Our subsequent tests aim to evaluate AdvIRL's ability to focus on objects within scenes. Additionally, these tests also assess the confidence levels achieved when using segmented images.

\subsubsection{Banana Scene:} Our initial tests, using the Segmentation-Reinforcement Learning pipeline, resulted in adversarial noise that misclassified the banana as a flatworm with confidences ranging between 35\% and 75\%. This demonstrates AdvIRL's effectiveness in applying adversarial noise to 3D-generated models across multiple viewpoints. Some images generated from this experiment are shown in Figure \ref{fig:AllExperiments}.

The noise generated in the banana scene demonstrates the high efficiency of our pipeline, as AdvIRL selectively applies noise to the banana alone, without impacting areas outside of the banana.

\subsubsection{Truck Scene:} We conducted a targeted attack with the cannon class as the target. AdvIRL successfully misclassified 15 out of 20 images as a cannon, with confidences ranging from 15\% to 70\%. This highlights AdvIRL's ability to produce effective targeted adversarial noise, posing a potential risk for vision models in applications like self-driving.

\begin{figure}[h]
    \centering
    \begin{subfigure}[b]{0.235\linewidth}
        \includegraphics[width=\textwidth]{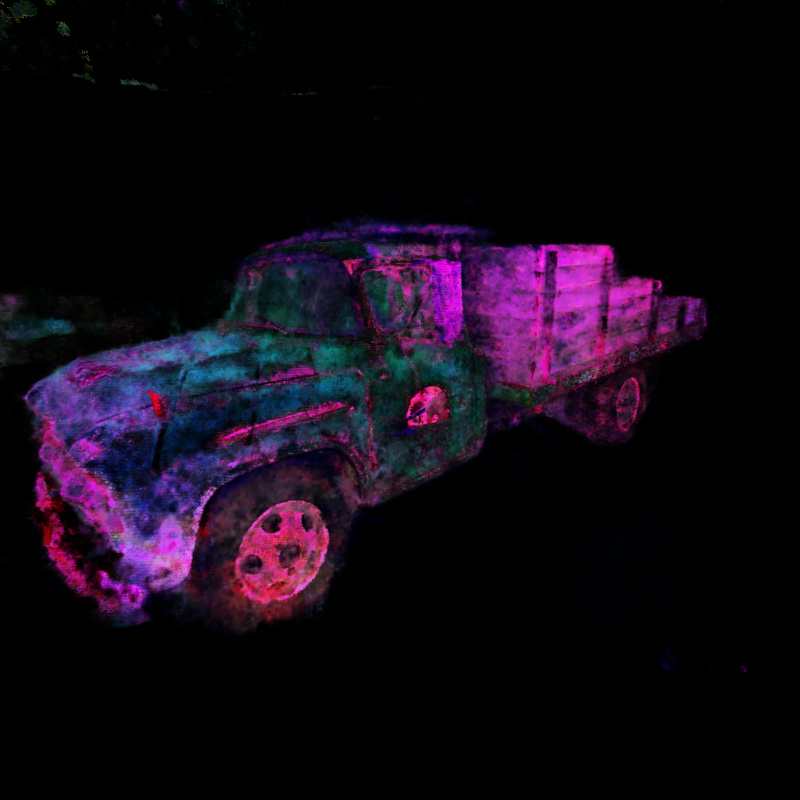}
        \caption{}
        \label{fig:truck_multi_angle_1}
    \end{subfigure}
    \begin{subfigure}[b]{0.235\linewidth}
        \includegraphics[width=\textwidth]{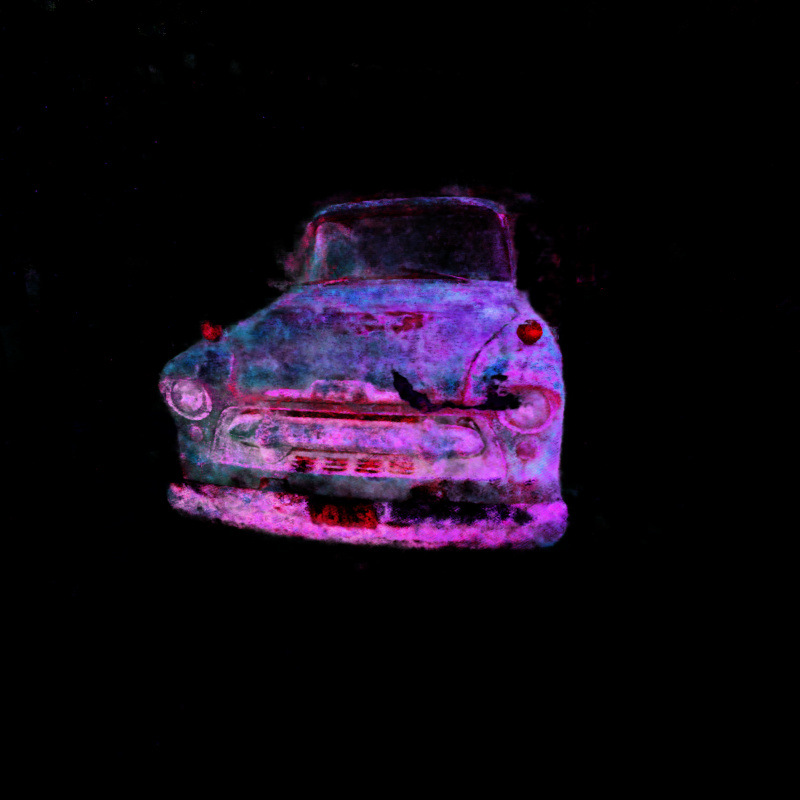}
        \caption{}
        \label{fig:truck_multi_angle_2}
    \end{subfigure}
    \begin{subfigure}[b]{0.235\linewidth}
        \includegraphics[width=\textwidth]{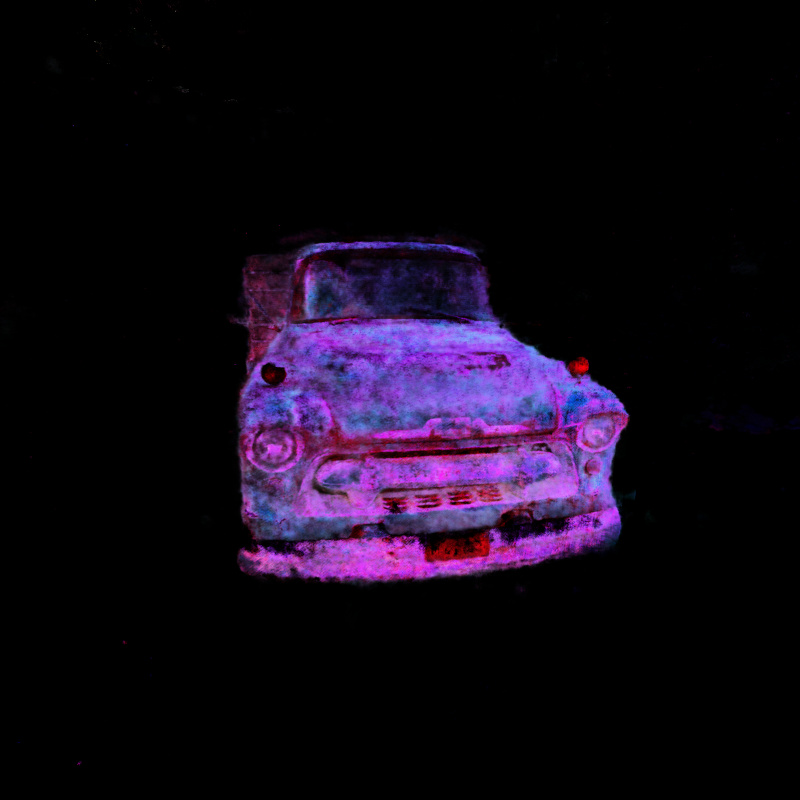}
        \caption{}
        \label{fig:truck_multi_angle_3}
    \end{subfigure}
    \begin{subfigure}[b]{0.235\linewidth}
        \includegraphics[width=\textwidth]{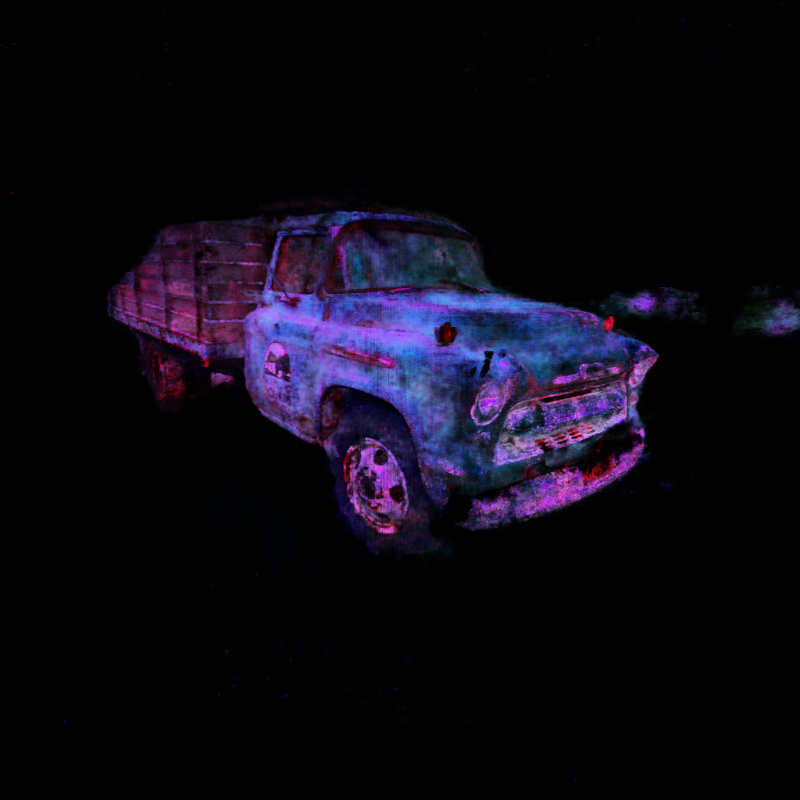}
        \caption{}
        \label{fig:truck_multi_angle_4}
    \end{subfigure}
    \caption{Images of the adversarially perturbed truck generated by AdvIRL from additional angles.}
    \label{fig:Multi_Angle_truck}
\end{figure}

\subsubsection{Horse Scene:} All 20 adversarial images were misclassified as other objects or animals (e.g., scorpion, Kerry Blue Terrier), with confidences ranging from 3\% to 70\%. Although some misclassifications had low confidence, AdvIRL demonstrated effectiveness in producing adversarial noise in a short time, which may be due to the complex shape of the horse. 

\begin{figure}[h]
    \centering
    \begin{subfigure}[b]{0.235\linewidth}
        \includegraphics[width=\textwidth]{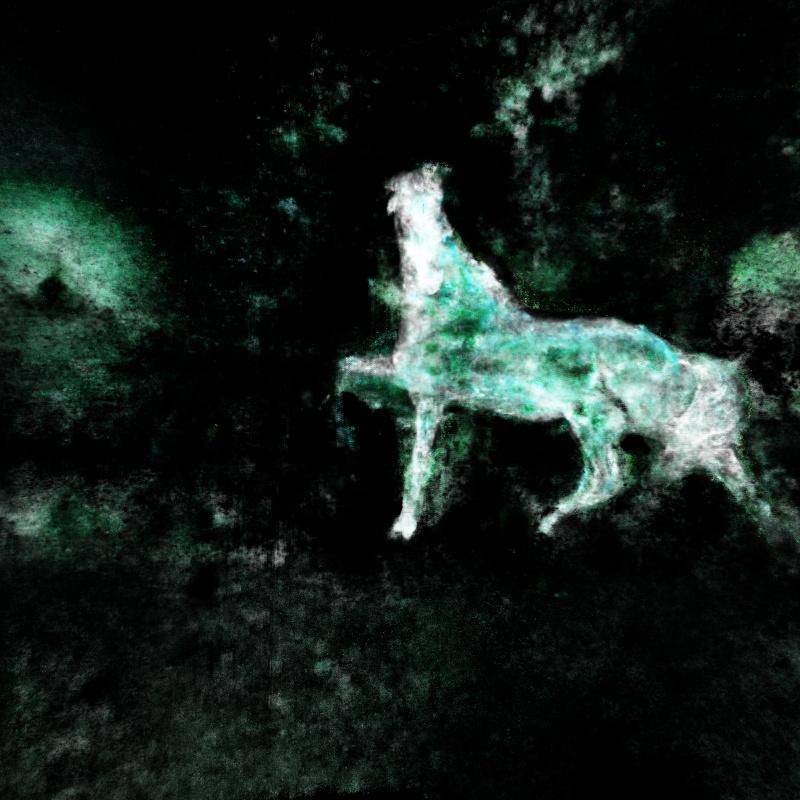}
        \caption{}
        \label{fig:horse_multi_angle_1}
    \end{subfigure}
    \begin{subfigure}[b]{0.235\linewidth}
        \includegraphics[width=\textwidth]{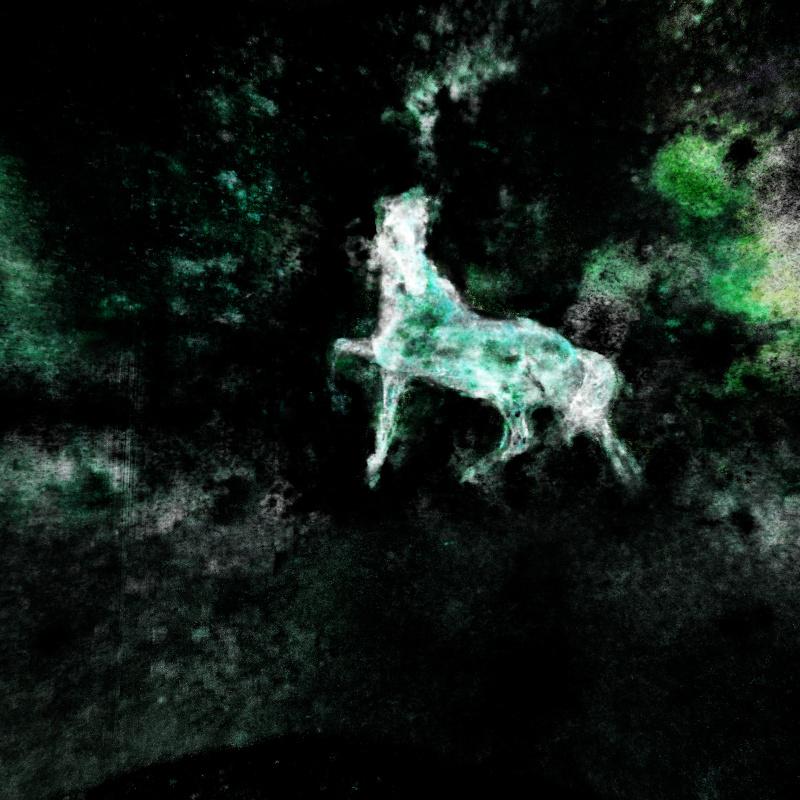}
        \caption{}
        \label{fig:horse_multi_angle_2}
    \end{subfigure}
    \begin{subfigure}[b]{0.235\linewidth}
        \includegraphics[width=\textwidth]{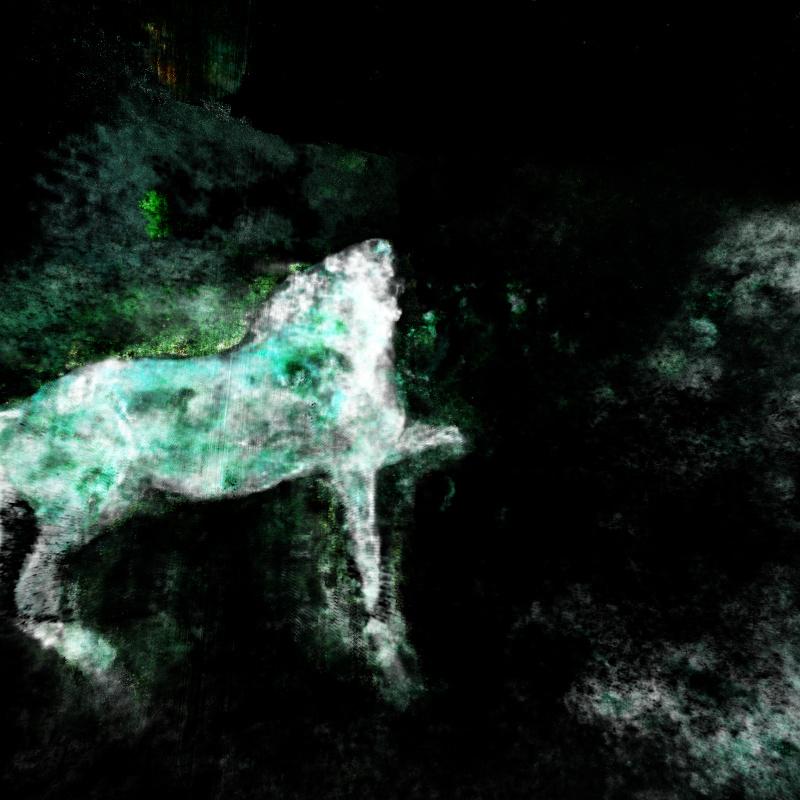}
        \caption{}
        \label{fig:horse_multi_angle_3}
    \end{subfigure}
    \begin{subfigure}[b]{0.235\linewidth}
        \includegraphics[width=\textwidth]{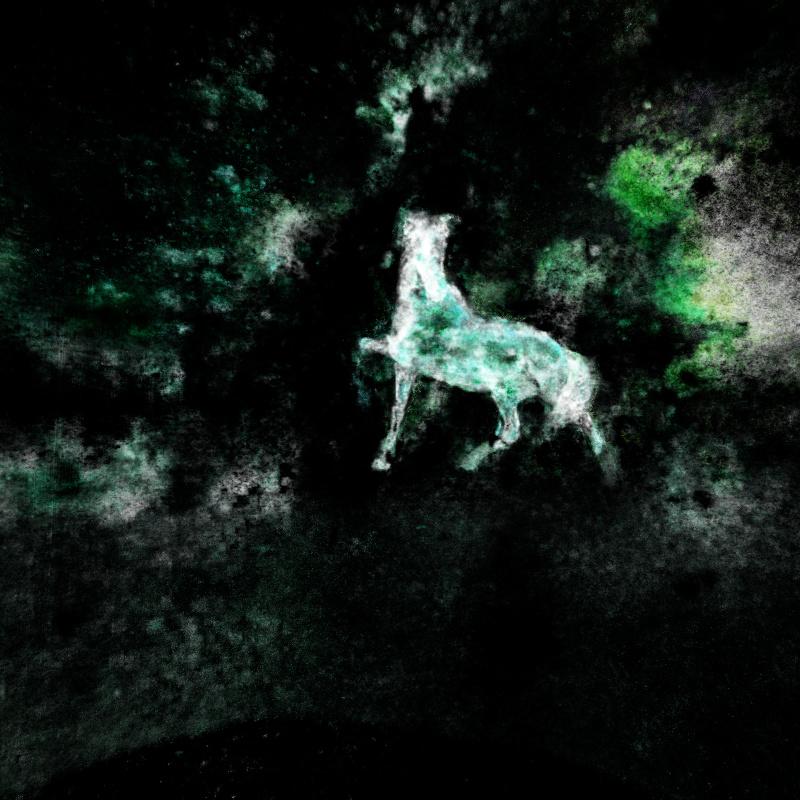}
        \caption{}
        \label{fig:horse_multi_angle_4}
    \end{subfigure}
    \caption{Targeted attack: 8 out of 20 horse images were misclassified as triceratops.}
    \label{fig:Multi_Angle_horse}
\end{figure}

In a targeted attack, where the target class was a triceratops, AdvIRL successfully misclassified 8 out of 20 images as triceratops, as shown in Figure \ref{fig:Multi_Angle_horse}. 

Interestingly, the targeted attack on the horse scene succeeded much faster than usual, taking less than a day to complete. This efficiency may be attributed to a modified reward system, where the number of images correctly classified as the target class (triceratops) was factored into the reward, improving AdvIRL's performance.

\section{Conclusion}
To our knowledge, AdvIRL is the first framework for conducting adversarial AI investigations on generated 3D objects of varying sizes and shapes in black-box scenarios. Central to our approach is the integration of an image segmentation model and a 3D rendering algorithm, such as NeRF, facilitating the generation of robust adversarial noise specifically targeting the texture of selected objects. Unlike conventional adversarial attacks, our model operates by adjusting the parameters of the NeRF algorithm, enabling it to render adversarial noise alongside images or 3D models.

AdvIRL stands out with its inherently black-box nature, relying solely on the input and output of the target model to generate effective adversarial noise. Unlike existing methods, which depend on knowledge of the target model's architecture or weights, AdvIRL requires no such information to produce robust and transferable adversarial noise. Despite the limited information available in a black-box setting, AdvIRL performs strongly in crafting adversarial noise for both targeted and untargeted scenarios.

Furthermore, our experiments employing the Segmentation-Reinforcement Learning pipeline (AdvIRL) underscore the efficacy of our approach in generating adversarial noise for both targeted and untargeted scenarios, providing a robust reward system that adequately incentivizes the agent to be implemented. These results suggest the capability of our model for adversarial noise generation and its potential usage in producing adversarial noise for 3D generative models, thus increasing the robustness of vision models against attacks.

\textbf{Future Works} While our algorithm performs well in producing both targeted and untargeted attacks, the act of producing targeted noise is highly dependent on the size of the action space. Since our action space represents the 13 million parameters of Instant-NGP, AdvIRL is given numerous potential modifications it can make. Due to this freedom, AdvIRL may take longer than if the action space were smaller. In turn, future works may focus on limiting the action space to a set of crucial parameters to modify, potentially leading to better and more efficient results. Furthermore, due to our usage of a pre-trained Detectron2 model, our segmentation is limited to the available classes in the model. Future work may involve incorporating Segment Anything Model 2 \cite{ravi2024sam2} for more generalized segmentation results.

\bibliography{thbib}

\end{document}